%% file: bare_jrnl_transmag.tex
\documentclass[journal,transmag]{IEEEtran}

\ifCLASSINFOpdf
\else
\fi

\usepackage{amsmath}
\usepackage{mathtools}
\usepackage{import}
\usepackage{times}

\usepackage{soul, color}
\usepackage{url}
\usepackage[hidelinks]{hyperref}
\usepackage[utf8]{inputenc}
\usepackage[small]{caption}
\usepackage{graphicx}
\usepackage{amsfonts}
\usepackage{booktabs}
\usepackage{import}
\usepackage{multirow, multicol}
\usepackage{subcaption}
\usepackage{array}
\usepackage{comment}
\usepackage{pifont}
\newcommand{\xmark}{\ding{55}}%
\newcommand{\cmark}{\ding{51}}%

\begin{document}
\title{A Survey on Artificial Intelligence for Music Generation: \\ Agents, Domains and Perspectives}

\author{\IEEEauthorblockN{Carlos Hernandez-Olivan\IEEEauthorrefmark{1},
Javier Hernandez-Olivan\IEEEauthorrefmark{1}, and
Jose R. Beltran\IEEEauthorrefmark{1}}
\IEEEauthorblockA{\IEEEauthorrefmark{1}Department of Electronic Engineering and Communications,
University of Zaragoza, Calle María de Luna 3 50018, Zaragoza}
\thanks{Manuscript received December 1, 2012; revised August 26, 2015. 
Corresponding author: M. Shell (email: http://www.michaelshell.org/contact.html).}}

\markboth{Preprint Submitted to IEEE}{Shell \MakeLowercase{\textit{et al.}}: Bare Demo of IEEEtran.cls for IEEE Transactions on Magnetics Journals}



\IEEEtitleabstractindextext{
\begin{abstract}
Music is one of the Gardner's intelligences in his theory of multiple intelligences. How humans perceive and understand music is still being studied and is crucial to develop artificial intelligence models that imitate such processes. Music generation with Artificial Intelligence is an emerging field that is gaining much attention in the recent years. In this paper, we describe how humans compose music and how new AI systems could imitate such process by comparing past and recent advances in the field with music composition techniques. To understand how AI models and algorithms generate music and the potential applications that might appear in the future, we explore, analyze and describe the agents that take part of the music generation process: the datasets, models, interfaces, the users and the generated music. We mention possible applications that might benefit from this field and we also propose new trends and future research directions that could be explored in the future.
\end{abstract}

\begin{IEEEkeywords}
Music generation, music information retrieval, artificial intelligence, deep learning, machine learning, human-computer interaction.
\end{IEEEkeywords}}

\maketitle

\IEEEdisplaynontitleabstractindextext
\IEEEpeerreviewmaketitle

\import{}{1-intro.tex}
\import{}{2-human-ai.tex}
\import{}{3-principles.tex}
\import{}{4-audio-symbolic.tex}
\import{}{5-datasets.tex}
\import{}{6-models.tex}
\import{}{7-hci.tex}
\import{}{8-eval.tex}
\import{}{9-further.tex}
\import{}{10-discuss.tex}
\import{}{11-concl.tex}

\section*{Acknowledgment}
This research has been partially supported by the Spanish Science, Innovation and University Ministry by the RTI2018- 096986-B-C31 contract and the Aragonese Government by the AffectiveLab-T60-20R project.




\ifCLASSOPTIONcaptionsoff
  \newpage
\fi

\bibliographystyle{IEEEtran}
\bibliography{references}








\end{document}

%% file: 1-intro.tex
\section{Introduction} \label{sec:intro}

\IEEEPARstart{M}{usic} generation is a research field of the Music Information Retrieval (MIR) that aims to generate new music. This field has gained much attention since recent deep learning (DL) models are being capable of generating longer coherent sequences. The evolution and release of new Deep Generative Models such as Dalle 2 by Open AI \cite{dalle} \cite{ramesh2022hierarchical} or Stable Diffusion by Stability.AI\footnote{\url{https://stability.ai/blog/stable-diffusion-public-release}, last access December 2022} in the field of image generation are attracting more users and are making not only the models themselves but also other parts of the generation such as human-computer interaction key parts of the development of AI-based tools.

There are many companies such as Sony, Google's Magenta project or Spotify and startups that are working on new music production and generation technologies, as well as music recommendation systems. A proof of this interest in music generation is the organization of workshops, talks and competitions such as the AI Song Contest that is held every year from 2020 and hosted 46 teams in 2022\footnote{\url{https://www.aisongcontest.com/}, accessed September 2022}.

To understand and compare music generation systems, it is necessary to understand how humans perceive, understand and compose music. Human understanding and perception of music depends on a variety of factors. The cultural background, the musical knowledge and the creativity of the composer are some of the main pieces that characterize the workflow of human musical composition. One way of composing is to develop an initial idea that the composer has. This idea contains not only symbolic or score-related musical principles, but also performance attributes such as timbre or dynamics, especially if the composer is an experienced musician. The continuation of the initial idea is guided by the decisions that the composer takes while composing a music piece \cite{levi1991field}. The decisions born from previous knowledge based on analysis and listening to many pieces of music acquired during the composer's career, i.e., the experience. This means that if we compare this process with AI-based music generation, there are more MIR fields involved in this task such as audio instrument classification or automatic chord recognition (ACR). Studying and analyzing other features related to musical principles might be necessary to build a complete AI music composition framework that we could name as a ``generalist music machine''. This generalist model could be based on Multimodal Deep Learning \cite{ngiam2011multimodal} and in previous generalist agents such as GATO \cite{gato}. In the music field, it has been proved that pre-trained models can help improving other tasks. This technique saves resources and training time and is a first step to a generalist symbolic music understanding model \cite{chou2021midibert}. Although the field is moving towards building deep learning models with a large number of learnable parameters, the musical language has defined rules that could be combined with such models to achieve this generalist model. As far as the generation of a piece of music is concerned, modeling the long-term structure remains an open area of research because, unlike text, music depends on two axes (pitch and time) that are dependent on each other \cite{review}. One approach to this would be the fusion of symbolic AI in which these rules can be defined and deep learning models that learn to combine these rules as humans do when composing music. Examples of those symbolic rules would be some music definitions such as the notes that belong to a certain chord, i.e., C, E and G in a C major triad chord.

The AI-based music generation process depends on factors other than those mentioned above, and one must take into account the biases that may appear in each of the agents in the AI-based composition workflow. There are similarities between human and AI composition processes that we will discuss in this article, however, there is a large gap between humans and machines when it comes to the understanding and perception of music \cite{review}.

In the musical field, the composition process or technique chosen will depend on the musical style we are working with \cite{collins2005synthesis}. While humans need some prior knowledge to be able to create a melody, AI needs to be trained on a large number of melodies in order to learn from them. However, when it comes to creating a complex composition, except in the case of genius musicians such as Mozart who had an innate ability for music, humans need a strong musical knowledge to develop the harmony, structure and instrumentation of a piece of music. Therefore, there might be some similarities in the musical composition process between a human and an AI. The difference lies in the different learning process and the creativity or extrapolation ability that humans have. Gardner's theory of multiple intelligences includes music as one of the 7 intelligences of a human. He defined the musical intelligence as the ability to produce and appreciate rhythm, pitch and timbre or the understanding of the forms of musical expressiveness \cite{gardner2011frames}. But where does the capability to compose music come from and where in the brain is this capability located? How can autonomous machine architectures learn and understand music to be creative as humans are? We will introduce these concepts in this paper and compare them to the AI models that have been proposed for music generation.

AI-based music generation can be studied from different perspectives \cite{briot2020deep}. The classification that Briot et al. proposed is based on five dimensions:
\begin{itemize}
    \item Input domain. Symbolic or audio.
    \item Model architecture. Sequence models, generative adversarial networks, etc.
    \item Generation. Autoregression, etc.
    \item Purpose. Melody generation, harmonization, improvisation, etc.
    \item Output's nature. Monophonic, polyphonic melodies with or without chords, multi-instrument music.
    \item Challenge. Human-computer interaction.
\end{itemize}

\begin{figure*}[!h]
    \centering
    \includegraphics[width=\textwidth]{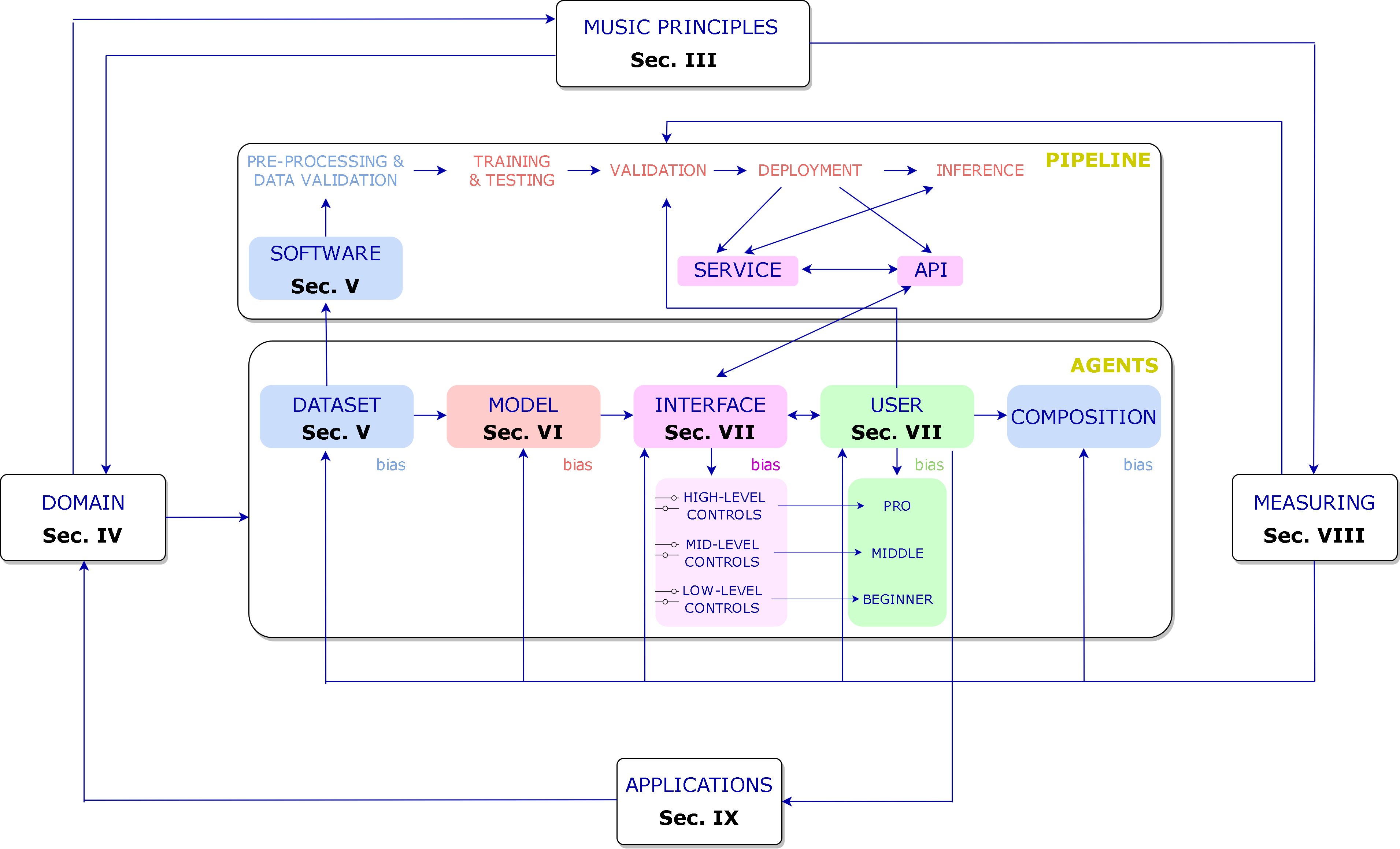}
    \caption{A general schema of the relationships between the agents in the AI-based composition workflow. The figure shows the different sections of this work in which each of the elements presented are addressed.}
    \label{fig:general}
\end{figure*}

In this paper, we will first outline the differences between human and AI-based composition processes. Then, we will present the agents involved in the music composition workflow such as the state-of-the-art datasets, models and algorithms that have been used until now to understand where we are, where this line of research is heading and the challenges that remain to be solved. Note that these agents can be extrapolated to other types of art generation with AI. In Fig. \ref{fig:general}, we introduce the most important parts of the AI-based music generation workflow. The workflow can be defined as follows: we select a domain (audio or symbolic music) that we want to focus on taking care and having some knowledge about the music principles that involve that domain. After the domain is selected, we can start describing the agents that we will discuss in each section of this paper starting from the data which we train the models to the end user who consumes the final products made with the music generation software. Note that each of the agents should be evaluated, however, in this paper we will only describe the models' evaluation that relies on the outputs they generate. When measuring the outputs we will need the music principles. Finally, the interaction with the users or the experience of the team that develops the technology can lead to new applications of music generation that could be in any domain.
In Fig. \ref{fig:general} we also show the pipeline that is developed in an industry application\footnote{We do not show all the elements of a real MLOps pipeline, we show the important parts that interact with the defined agents to show the general concepts.}

\subsection{Paper Organization}
This paper is organized based on the agents that are part of the music generation systems as follows: in Section \ref{sec:human-ai} we compare both human and AI-based music generation models, in Section \ref{sec:principles} we introduce the music principles, in Section \ref{sec:audio_symb} we introduce and compare the two big domains in music generation: audio and symbolic music, in Section \ref{sec:datasets} we give an overview of the datasets used in this field for both audio and symbolic domains, in Section \ref{sec:models} we present the models and architectures that are being used to build music generation technologies, in Section \ref{sec:hci} we discuss and introduce the human-computer interaction works that have been proposed in this field, in Section \ref{sec:eval} we present how these models are being evaluated from both subjective and objective perspectives, in Section \ref{sec:future} we present further works and studies that are gaining attention in this field, in Section \ref{sec:discussion} we discuss the past, present and future of AI-based music generation, and finally in Section \ref{sec:conclusions} we present the conclusions.

%% file: 2-human-ai.tex
\section{Human vs AI-based Music Generation} \label{sec:human-ai}
Understanding the human's brain is important to develop AI tools that could imitate cognitive process \cite{neuroai}. The field that aims to study such processes to feed and improve AI models is called NeuroAI. A better understanding of the human brain could help building creative AI tools that would be able to extrapolate and be creative as humans are. A big question about AI-based models is their capability to be creative, but what does creativity mean? How can we measure it and from where does it come from? In this section we introduce the concept of creativity and we give an overview of how humans understand and create music.

\subsection{The concept of ``creativity''}
Amabile defines \textit{creativity} as a \textit{``process which results in a novel work that is accepted as tenable or useful or satisfying by a group at some point in time''} \cite{amabile2018creativity}. 
The concept of creativity can be seen from three different points of view: the \textit{product}, the \textit{identity} which produces the product and the \textit{process}. The consensual definition of creativity given by Amabile says that: ``\textit{...creativity can be regarded as the quality of products or responses judged to be creative by appropiate observers, and it can also be regarded as the process which something so judged is produced}''. This definition is based on the final product, not on the process or the identity that produces such creativity. These definitions, in musical terms, mean that new music can be considered a creative process if it is accepted by a group of people, i.e., accepted by a part of our society. From these definitions we can argue that AI-based music could be considered creative if society accepts it and finds it useful for whatever purposes.
When it comes to the art made by machines, we can also define the concept of \textit{computational creativity} which is defined by Colton and Wiggings \cite{coltonwiggings} as ``\textit{the science, engineering and philosophy of computational systems which, by taking on particular responsibilities, exhibit behaviours that unbiased observers would deem to be creative}''. More specifically, in the field of MIR, there is a subfield of music generation that aims to study and computationally simulate machine creativity \cite{agres2016evaluation}. This field is called Music Metacreation (MuMe).

\subsection{Measuring Creativity}
Researchers have attempted to measure creativity with three different techniques: by making an objective or subjective analysis and by using creativity tests \cite{amabile2018creativity}. In the field of AI-based music generation or composition, the most commonly used technique to measure the quality of a generated composition is the Turing test or a survey with questions that are somewhat close to creativity tests. Amabile mentions that, because of the narrow ranges of abilities that can be assessed in a creativity test, it is inappropriate to assume that a test is a general indication of creativity. Ada Lovelace proposed some questions to measure machines creativity that Boden summed into the Lovelace questions \cite{boden2004creative} that turned into the Lovelace Test (LT) for creativity. The LT is harder to pass than the Turing Test since the algorithm (the interrogated) must fool the programmer (the interrogator). In the AI research field that aims to generate new music, usually the results are measured with subjective tests, which is not always reliable because the results depends on the population's culture and background \cite{subjective}. Some objective evaluation measures have been proposed that aim to fill the gap that subjective evaluation cannot fill, which we mentioned in Section \ref{sec:eval}, but there is still a big room of improvement in this area \cite{loughran2016generative}.

To try to find the parts of the brain where this creativity is found and how music can be represented as a language similar to written language, we will now describe the relationships between written and musical languages.

\subsection{Music as a language form}
In spite that some musicologists do not consider music as a language due to its subjectivity, others agree that there is a connection between the structural elements of music and other languages.
Sloboda claims that ``\textit{musical language can be understood, on a structural level, like any other language}'' \cite{sloboda1990music}. Music evokes emotions through the music elements such as the harmony, rhythm or timbre \cite{cooke1959language}. If we consider that a note or a property of a note is a character or a word, and extend this relationship to other music elements, we could encode music as text and apply Natural Language Processing (NLP) techniques to music. By establishing a relationship between the two structures, we could apply existing models in language generation to music generation.

The area of the brain responsible for language comprehension and interpretation is Wernicke's area (or posterior speech area), which is located in the left hemisphere. It has been shown that the brain spontaneously participates in predicting the next word before it occurs without the need for explicit instruction \cite{nature2022}. This is precisely what deep language autoregressive models (DLMs) do, thus suggesting that the predictions of the following word from DLMs and humans are similar in natural contexts \cite{nature2022}. On the other hand, different music-related functions are also attributed to this part of the brain according to Platel et al. \cite{platelH1997}, such as the recognition of rhythmic, temporal and sequential structures, and is related to semantic representations of musical stimuli (melody recognition and identification) \cite{andrade2003brain}. However, the right temporal cortex is involved in the processing, recognition and discrimination of timbre and pitch \cite{zatorre1994}. Lu et al. \cite{lu2015brain} used a functional magnetic resonance imaging (fMRI) to explore the functional networks in 17 professional composers during the creation of music. This allowed to demonstrate that there is a functional connectivity between the anterior circulate cortex, the right angular gyrus and the bilateral superior frontal gyrus during the composition process, which means that in spite that the primary visual and motor areas do not play an important role when composing music, the neurons in these areas increase the functional connectivity between the anterior cingulate cortex and the default mode network to plan the input of musical notes and emotions. 

Thus, viewing music as a form of language and discarding attributes of interpretation such as timbre, we can argue that the parts of the brain responsible for understanding and knowing how to interpret the structure of a text and a piece of music are the same. However, maths and the probability theory on which AI models are based are different from the way a human brain works and processes information.

\subsection{Human Composition and AI-based Music Generation Processes}
The human composition process involves cognitive processes, such as attention to action, response generation, action planning and monitoring, and inhibition of repetitive responses \cite{lu2015brain}. LeCun \cite{lecun2022path} describes a system architecture for autonomous intelligence or \textit{word model}. The usage of neural networks to model this complex systems was introduced in the 1980s in works by Jordan et al. \cite{jordan1988supervised}, Widrow et al. \cite{nguyen1990truck}, etc. This \textit{world model} has the following modules: The configuration module that takes inputs, the perception module which estimates the current state of the world, the world model module that predicts possible future world states, the cost module which computes the output or ``energy'', the short-term memory module which remembers the current and predicted world states and the actor module that computes proposals for action sequences. Although this is a theoretical ``world'' model, it could be built a general music machine that would be capable of understanding different musical representations (audio or scores), analyzing them, creating new music and interacting with the created music. 

\begin{figure*}
    \centering
    \includegraphics[width=1.6\columnwidth]{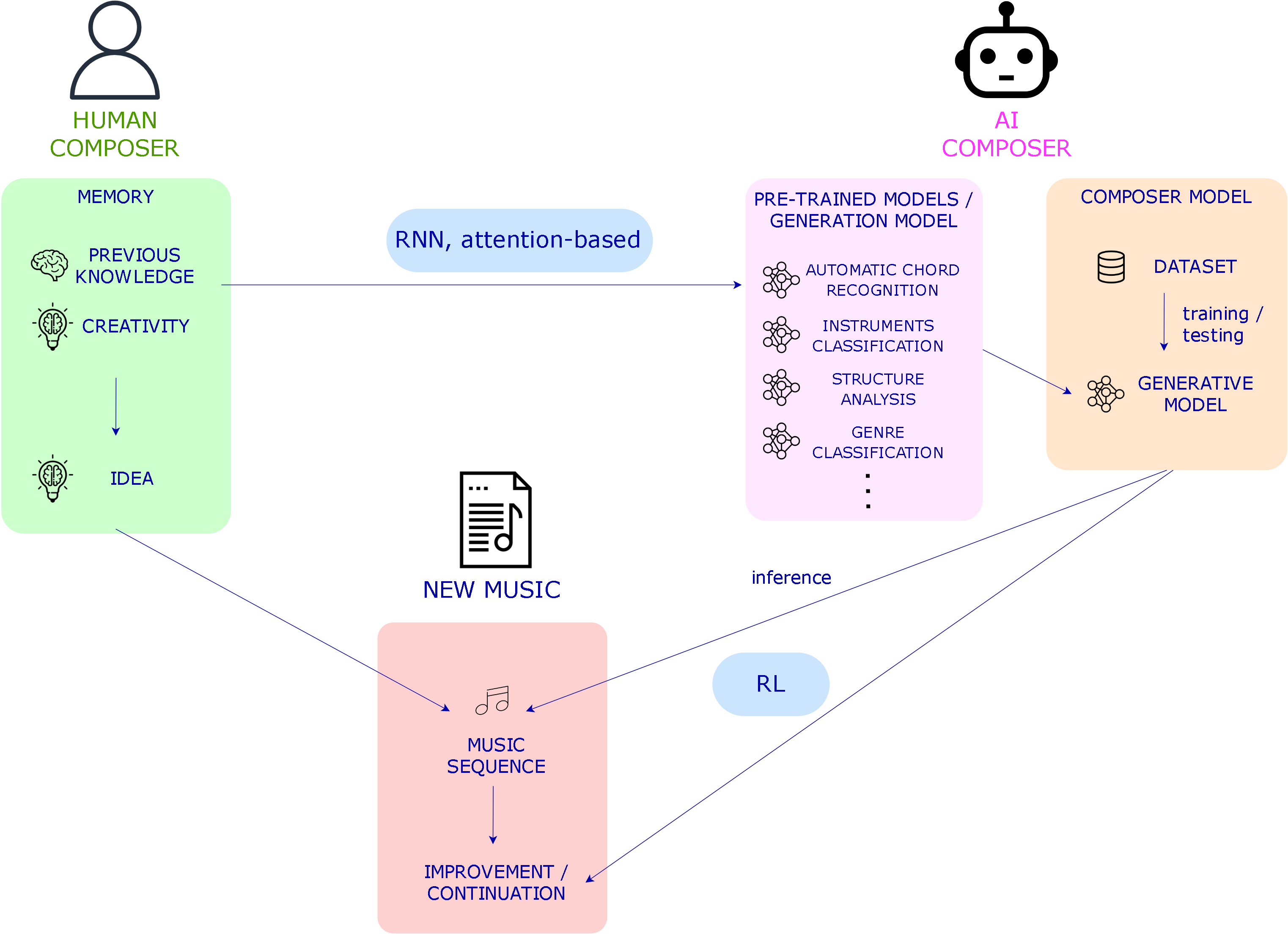}
    \caption{Comparison between AI and human composition processes. On the right, the human composer makes music with the previous knowledge that he or she has, and also with the creativity that leads to music ideas that can be rhythms, notes, etc, or their combinations in motifs or melodies. On the left, the AI composer makes music with models trained on different MIR tasks (the ``previous knowledge'') combined with a model that composes music being conditioned. The pre-trained models that represent the previous knowledge in the AI composer could be replaced in the future by a generalist model that would understand music, and the memory could be comparable to a large RNN. Below, we see the music piece (or the ``world'' if we consider all the existing music) that is created or completed by the human or the AI composer, in the case of the latter with Reinforcement Learning techniques, sequence models, etc.}
    \label{fig:processes}
\end{figure*}

In Fig. \ref{fig:processes} we show a scheme of both human and AI-based music composition processes where both the human and the machine interact with the ``world'' or piece of music by performing actions thanks to their memory, something like large Recurrent Neural Networks (RNNs) with more memory or Transformers, or prior knowledge which in AI-based models can be replaced by pre-trained models that could feed the machine or composer model (or a multitasking model that could learn to analyze the music to then compose it). This idea of having task-specific pre-trained models in other MIR tasks would allow conditioning the generation of various ways or levels. However, having a generalist pre-trained model for symbolic music that can be fine tuned to perform different MIR tasks has proven to be an alternative to developing task-specific models. \cite{chou2021midibert}. There might be other alternatives for modeling a generalist model since recent advances in Automatic Music Transcription (AMT) could also be used as generative models, i.e., MT3 \cite{mt3} or DiffRoll \cite{diffroll}, a Transformer-based and a diffusion-based model, respectively. In either case, the output music can be created from scratch, painted or continued from a given sequence.

\subsection{Agents in AI-based Music Generation}
Now, diving into the AI part of music generation we find 5 agents or entities in a music composition framework: the training dataset, the model, the interface and the user, and the final composition that needs to be measured, that will be presented in sections \ref{sec:datasets}, \ref{sec:models}, \ref{sec:hci}, \ref{sec:eval}, respectively. In Fig. \ref{fig:general} we can see a general scheme of the agents or entities that are part of a AI-based music generation or composition process. Each of the agents that may be present in a music generation job is susceptible to bias. While datasets and models may have objective biases due to the nature of the input music or the model architecture, interfaces and users have subjective biases due to the subjectivity of the music. In this paper, we will not make any bias analysis.

We can also define two main processes related to these agents: training or development and measurement. While training occurs between the database, the model and/or the user interface, the measurement or evaluation stage occurs in all agents, and to a greater extent between the user and the interface. If the proposed model for music generation does not address the end user or the interface, we can group the training and measurement stages only at the database and the model.

%% file: 3-principles.tex
\section{The Music Principles} \label{sec:principles}
So far, we have covered the human and AI-based music composition but, what is it and where does the previous knowledge for composing music come from? In this section we will describe the music principles that are the pillars of music. This principles contain the definitions of the basic elements that one needs to understand before composing music.

Music can be seen as a hierarchical structured language \cite{review}. The hierarchy is related to the music principles which are the pillars of music \cite{walton2005basic}.
The musical principles are: harmony, rhythm, form or structure, melody, timbre and dynamics. 
These principles can be analyzed in the symbolic domain and the audio domain. While harmony, rhythm, structure and melody are more related to the symbolic domain, timbre, effects or dynamics are part of the interpretation, i.e. the audio domain. It is also worth mentioning that melody can be considered both in the perspective of the symbolic domain and the audio domain if we take into account how it is interpreted. Texture is a consequence of how we combine these principles. Apart from the musical principles, there are composition techniques that composers use to combine these principles. Examples of those processes are the instrumentation or the orchestration which we will discuss later in this section.
The definition of the music principles or parameters \cite{christensen2006cambridge} are:
\begin{itemize}
    \item Harmony. It is the superposition of notes in intervals that form chords. The highest-level in harmony is the key or tonality in tonal music. The chords that compose a section usually belong to the same tonality.
    \item Music Form or Structure. It is the highest level in the time dimension. Then, sections are organized in concatenations of music phrases that are developments of small music units that are the motifs.
    \item Melody. Melodies can be monophonic or polyphonic depending on the notes that are played at the same time division, and homophonic or heterophonic depending on the melody accompaniment. 
    \item Timbre. The quality of sound that makes musical instruments or voices sound different from each other when they are playing the same frequency with the same intensity.
\end{itemize}

Christensen \cite{christensen2006cambridge} also adds the meter, the counterpoint and genre or style, and excludes the timbre from the music principles. However, we classify the counterpoint as a composition technique and we add the timbre since it is an important part of the composition, in particular, in the performance, that would be important in future audio generation works.
Instrumentation and orchestration are music techniques that organize and arrange the music content. Instrumentation is the selection of instruments that are present in a music piece and orchestration is the technique of assigning melodic parts or accompaniments to each instrument depending on the dynamics and colour that the composer aims to achieve. In symbolic music software, instruments are organized as \textit{tracks} which contains the information and notes of each instrument, as happens in a music sheet.

So, from the AI perspective, should we call the field of AI music creation as music \textit{generation} or \textit{composition}? This depends on the goal of each work. If we want to build models that generate music and do not care about how the music is being generated in terms of musical principles and music theory, we could call the problem music generation. But if we care about these principles and intend to build interfaces that allow users to interact with the model at various levels related to musical principles to have more control over them, and if users can learn music with these models, we can call this field music composition.

\subsection{Human music composition}
As we mentioned in Section \ref{sec:human-ai}, when a composer creates a piece of music lots of decisions are taken. This occurs in all musical genres or styles, even in improvised musical pieces in genres such as jazz, where decisions are made while playing. In Western classical music it is common to compose a short musical sequence called a \textit{motif}. The motif contains the musical ideas that will be developed in a musical section. These ideas contain rhythmic and harmonic information that can be arranged in different ways creating melodies with their corresponding accompaniments. Another way of composing music can be the creation of a melody that is then harmonized, or also the construction of a harmonic progression from which a melody will be built. These ways of composing music by human beings depend on the composer and the musical genre. 

After a melody, chord progression or motif is created, the composer completes a music phrase that contains similar music ideas. Music phrases are concatenated with \textit{bridges} that are transitions from one phrase to the next one. A group of music phrases form a section which usually finishes with a \textit{cadence} or transition to the next section. There are several ways of structuring music as happens with the Sonata form in the Western Classical music \cite{drake2000beethoven}, however, each composer adapts the structure to he or her needs. In the next sections, we will see how these ways of composing have been modeled by AI systems to mimic these processes. However, AI itself can change the way humans compose music, i.e., for example, with sound exploration in latent spaces.

In terms of the representation of music, it has two dimensions, the time and the harmony. Both dimensions depend on each other. The musical information in the time dimension and in the symbolic domain is grouped by measures containing the notes. This allows to group the musical information with a time unit called time signature.
Of all the elements that make up musical language, notes are at the lowest level. The measures, musical phrases or melodies can be represented at the next higher level and the structure of the piece at the highest level.

\begin{figure*}
    \centering
    \includegraphics[width=0.8\textwidth]{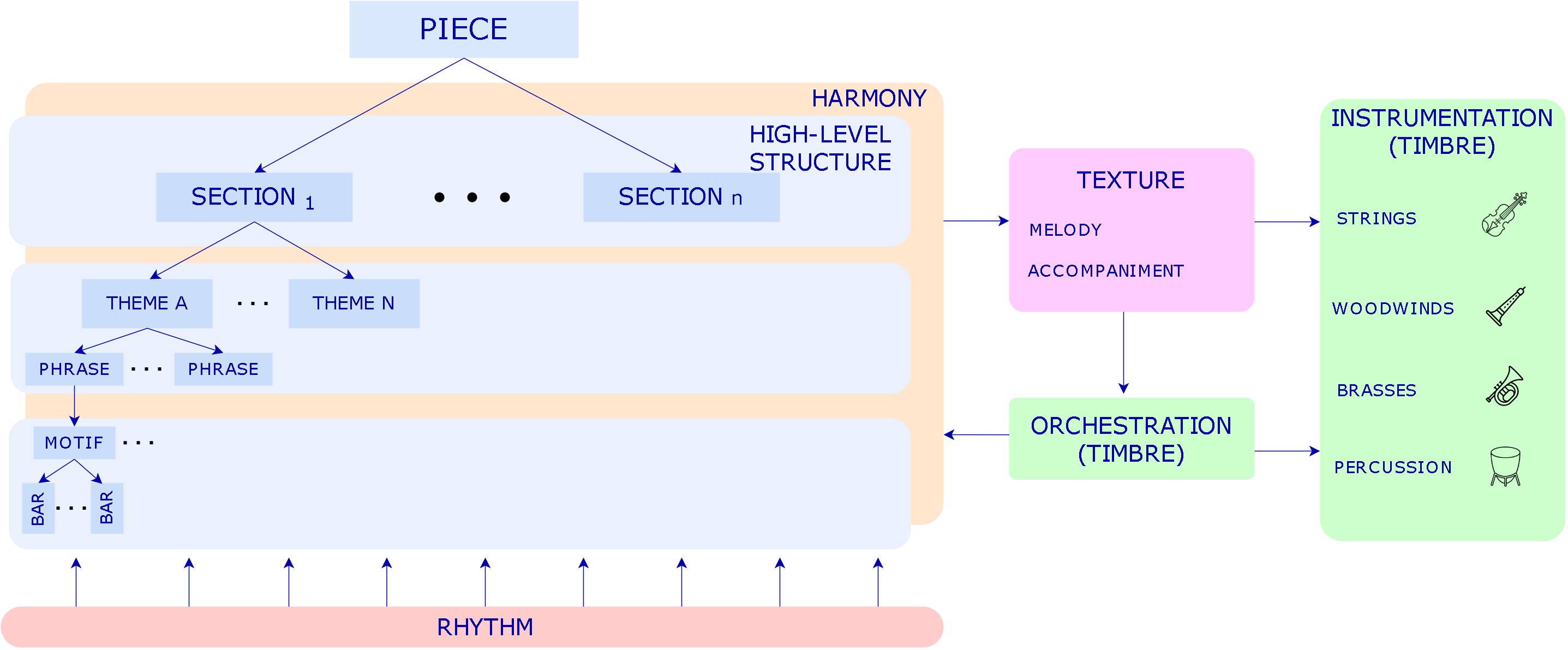}
    \caption{The music principles and their relationship. The music piece is structured in different levels from sections to motifs (this depends on the music genre). The harmony and rhythm are the two axes in music that complement the music form. From them, the melodies, accompaniments can be derived or constructed with counterpoint techniques and arranged with the instrumentation and/or the orchestration.}
    \label{fig:structure}
\end{figure*}

\begin{figure*}
    \centering
    \includegraphics[width=0.7\textwidth]{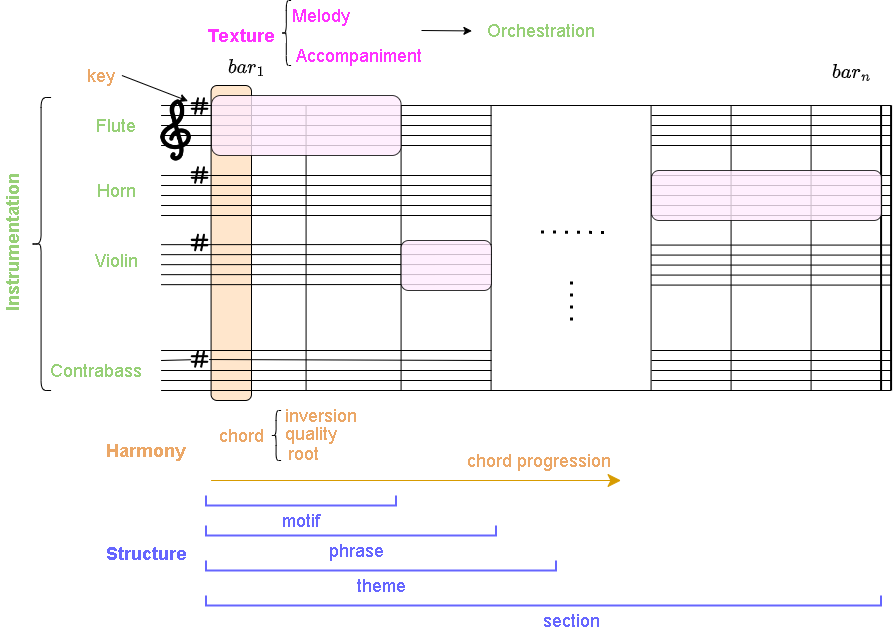}
    \caption{Music score schema with the music principles. We show how musical principles are related to each other in a score-like representation. Note that a music sheet does not normally have all the harmonic or structure annotations such as the chords or the melody identification but it does have the instruments, the key, or the dynamics of each instrument to tell the musicians or conductor how the piece should be played.}
    \label{fig:melody}
\end{figure*}

In Fig. \ref{fig:structure} we show the formal structure of a generic music piece with the music principals as its building blocks, and Figure \ref{fig:melody} represents a schema of the music principles in a music sheet representation.

%% file: 4-audio-symbolic.tex
\begin{figure}[h]
    \centering
    \includegraphics[width=\columnwidth]{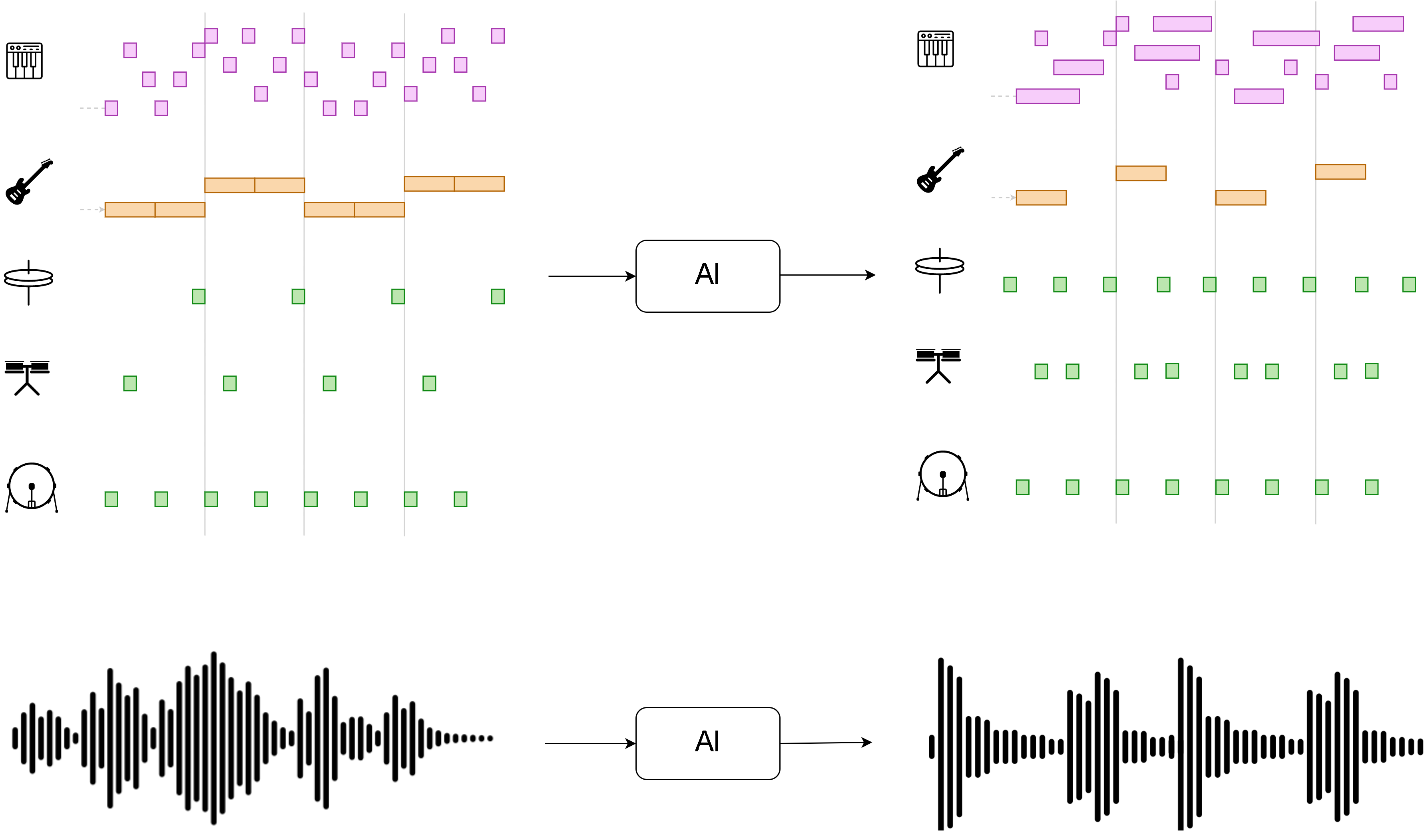}
    \caption{Symbolic pianoroll (up) and audio waveform (down) domains.}
    \label{fig:comparison}
\end{figure}

\section{Symbolic and Audio Domains} \label{sec:audio_symb}
After describing the music principles in the previous section, we will now dive into the AI-based music generation that will be the main topic from this section to the end of the paper. In this section we will describe the domains of music generation attending to the input and output signals nature. AI-based music generation can be seen from two perspectives according to the input and output nature of the data: the symbolic and the audio domains. We will describe both domains and the most-common music representations that have been used in the state-of-the-art. In Fig. \ref{fig:comparison} we show the schema of an input and output-like symbolic and audio music data.

\subsection{Symbolic Music Generation}
Symbolic models work with music information at different levels such as the notes duration (or the note on and note off events in the case of MIDI files), with pitch and dynamic attributes, etc. However, the output music would not have performance attributes due to the fact that the input MIDI files might not contain that information. For the perspective of building a model that will be used by a user and depending on the application we would like to build, the output music should be synthesized with performance attributes which can change the way the music is perceived. It is easier to retrieve features related to the music principles in this domain due to the fact that we already know the notes information. We can detect chords or keys with probabilistic algorithms and with recent neural network models for harmonic recognition \cite{chen2021attend}, detect the structure or boundaries \cite{boundaries2021hernandezolivan} of the pieces or note sequences with the well-known Self-Similarity Matrices (SSMs) \cite{foote1999visualizing}, and we can use this information to encode the discrete music information to train and condition an AI-based model.

Symbolic music generation systems can compose music \textit{from scratch} or {by conditioning}. It depends on the model the way we want it to compose music. In sequence models, both options are suitable since we can feed previous tokens to the network.

\subsection{Symbolic Music and Performance}
One of the weaknesses of generating symbolic music is the lack of performance attributes that these representations have. As we mentioned in Section \ref{sec:human-ai}, a complete music piece contains not only symbolic but also performance attributes which are a key part of how music is played and perceived. Apart from classical techniques for synthesizing music, there have been emerging new models that use deep learning techniques for symbolic music synthesis with controllable attributes. One example of such models is the MIDI-DDSP \cite{wu2022mididdsp}. This model provides a controllable synthesis where the symbolic music is converted into audio with specific playing techniques such as vibrato, staccato, etc. The integration of these models along with symbolic generation would provide an entire framework for co-composing music with Artificial Intelligence.

\subsection{Audio Generation}
Audio music generation consists of generating audio signals instead of symbolic musical representations. This allows training models with raw audio as inputs. The advantage is that we do not need symbolic music which sometimes may not be easy to find, and that the output does not need a post-synthesis to be reproduced. On the one hand this is an advantage, but on the other hand, raw audio is more difficult to control with performance attributes than symbolic music synthesis. Despite this, there are techniques such as timbre transfer of effects modeling with deep learning that could improve, condition and allow to control the generation of the output waveform. Some models that generate raw music are models originally designed for speech, but demonstrate good results in the field of music generation.

The most-common architectures for audio generation are VAEs and GANs. One of the most well-known models for generating raw audio is Wavenet introduced in 2016 by DeepMind \cite{wavenet}. The model uses dilated convolutions to allow the receptive field to grow exponentially with depth and cover thousands of timesteps that are composed by a sequence of samples which, in a raw signal, can be 16,000 samples per second or more.

\subsection{Style Transfer} This is a case of \textit{neural style transfer} that is a extended technique in arts generation. In the music field this could serve to two different purposes: \textit{timbre transfer}, \textit{genre transfer} and \textit{performance transfer} \cite{dai2018music}. The first one consists on changing the timbre of an instrument by preserving its time-frequency properties. This task is not closely related to music generation, but with the advent of new interfaces for music generation that we will discuss in Section \ref{sec:hci}, it might be a technology that could be added to a music generation tool to help users to choose the right sounds for their compositions. The second one consist on changing the genre or style of a composition which alters the symbolic information of the music. The third one applies to control the performance. This means that we can alter the performance attributes of an instrument such as its dynamics or playing techniques and not only its timbre. An example of a state-of-the-art model for timbre transfer is the DDSP \cite{engel2020ddsp}.

\subsection{Input Representations}
Whereas in the symbolic music we use formats that store the value of the notes or \textit{pitch}, their duration, dynamics and other data at the piece-level such as the instruments, key or tempo, in the audio domain we normally have the raw signal. Depending on the sources from which we obtain these files, we can find annotated data in audio datasets and also in symbolic music such as the high-level structure labelling which is difficult to find in symbolic music formats. The most common symbolic music formats are MIDI files, MusicXML and ABC notation. From these files and thanks to the information they provide we can create music encodings or tokenizations to train sequence models, construct one-hot vectors or pianorolls. In audio, it is common to use spectrograms, or any time-frequency representation as inputs, provided they are invertible, so that they can be reconverted back to audio.

%% file: 5-datasets.tex
\section{Datasets} \label{sec:datasets}
We described the domains of the music generation field in the previous section. So far, having described the music principles and the domains and following the workflow showed in Fig. \ref{fig:general} we will now dive into the first agent of the music generation: the datasets and the software that allow to process them. From the data perspective we find some questions that we will discuss in section \ref{sec:discussion}: do the current datasets for music generation represent all kinds of music genres? What annotations would they have to contain to be more useful in what concerns to controllability?

\subsection{Datasets for Music Generation Research}
In the history of the field of music generation, a few datasets have become popular in the research community. In spite of that, there has not been a comprehensive analysis of these datasets and the biases they may contain, and therefore the model's output is the main thing that has been studied when evaluating music generation systems.
Some of the most-commonly used datasets for symbolic music generation are: the Lakh MIDI dataset (LMD) \cite{raffel2016learning} which contains MIDI files aligned with the Million Song dataset \cite{million_song}, the Bach Chorales dataset (JSB Chorales) \cite{LewandowskiBV12}, the MAESTRO dataset \cite{maestro} and the Meta-MIDI dataset (Meta-MIDI) \cite{metamidi}. Some of these datasets contain multi-track music whether others are specifically designed for a concrete music genre. With the advent of new deep generative models, new datasets are being created such as the Symphonies dataset that proposed in SymphonyNet \cite{symphonynet}. More datasets such as POP909 \cite{wang2020pop909} cover other tasks in the music generation field such as the accompaniment generation. In addition, datasets that are commonly used in other domains such as Slakh2100 \cite{slakh2100} for Automatic Music Transcription \cite{hernandez2021comparison} could be used in symbolic and audio music generation since some of them contain symbolic and audio files.
In Table \ref{tab:datasets_analysis} we provide a summary of the features that each dataset contains.

In the audio domain, the main advantage is that we do not need the symbolic data to train the model, which sometimes is difficult to find. We can train our model directly with the audio signal, which opens more possibilities to the field of music generation, since it is usually easier to find audio files than MIDI files. However, the copyright that protects audio and MIDI files is still an open problem in the MIR community that leads researchers use a very few datasets in the music generation task. In this field, some models are trained with MAESTRO audio samples and others with music extracted from the Internet.

Both audio and MIDI files in the datasets are usually protected by a copyright license for commercial purposes. There are studies that seek to define who owns the music, whether it is the team that builds the model, the proprietary dataset, or the user who interacts with the model and creates the final piece. We give further details of the ownership in AI-based music generation systems in Section \ref{sec:future}.

\begin{table}[!t]
\caption{Common music generation datasets.}
    \centering
    \begin{tabular}{
    p{2.6cm}  
    >{\centering\arraybackslash}p{0.3cm}
    >{\centering\arraybackslash}p{0.5cm}
    >{\centering\arraybackslash}p{0.9cm}
    >{\centering\arraybackslash}p{1.2cm}
    >{\centering\arraybackslash}p{1.2cm}
    }
    \toprule
        Dataset & MIDI & Audio & n files & Genres & Instr.\\
        \midrule
        MAESTRO \cite{maestro} & \cmark & \cmark & 1,200 & Classical & Piano\\
        LMD \cite{raffel2016learning} & \cmark & \xmark & 176K & Pop, Rock & 128 MIDI\\
        JSBC \cite{LewandowskiBV12} & \cmark & \xmark & 500 & Chorales & 4 voices\\
        JSBCF \cite{jsbfakes} & \cmark & \xmark & 382 & Chorales & 4 voices\\
        Meta-MIDI \cite{metamidi} & \cmark & \cmark & 436,631 & all & 128 MIDI\\
        Symphonies \cite{symphonynet} & \cmark & \xmark & 46,359 & classical & multi\\
        GiantMIDI-Piano \cite{kong2020giantmidi} & \cmark & \xmark & 10,855 & classical & piano\\
        NES-MDB \cite{donahue2018nes} & \cmark & \xmark & 5,278 & games & multi\\
         \bottomrule
    \end{tabular}
    \label{tab:datasets_analysis}
\end{table}

\subsection{Software for Music and Audio Processing}
We can make use of some Open Source software to pre-process, extract features from music, train DL models and evaluate them in both symbolic and audio domains.
On the one hand, in the symbolic domain we can use \texttt{musicaiz} \cite{musicaiz} to generate, analyze, visualize and evaluate symbolic music, \texttt{jSymbolic} \cite{jsymbolic} for analysis purposes, \texttt{music21} \cite{music21} to analyze symbolic music in all its forms and formats, \texttt{Humdrum} \cite{humdrum}, \texttt{mido} for low-level MIDI processing, \texttt{pretty\_midi} \cite{pretty_midi} to process MIDI files, \texttt{muspy} \cite{muspy} to prepare dataloaders to train DL models for music generation, \texttt{miditok} \cite{miditok} to tokenize MIDI files with state-of-the-art encodings that we will introduce later in this section, or \texttt{note\_seq} to process symbolic music and use it to train DL models (the library is maintained by Google's Magenta team). On the other hand, if we work in the audio domain, some of the most-used software are \texttt{librosa} \cite{librosa}. \texttt{Essentia} \cite{essentia} or \texttt{madmom} \cite{madmom} which are also a well-known software for building audio applications in spite of the copyright license they have for commercial applications.
The commonly used software to measure music in the MIR community is \texttt{mir\_eval} \cite{mir_eval}. In Table \ref{tab:softwares} we show a comparison between these software libraries.
There is still a need of more open source frameworks for preserving and adopting good practices in the MIR community to make the research results available and reproducible. It is worth to mention the Magenta\footnote{\url{https://github.com/magenta/magenta}, Accessed October 2022} and the Muzic\footnote{\url{https://github.com/microsoft/muzic}, Accessed October 2022} projects by Google and Microsoft respectively, which contain all the models that their respective teams have been proposed in the past years.

\begin{table}[!h]
\caption{Frameworks for MIDI and symbolic music generation, analysis and representation.}
\begin{tabular}{
    >{\centering\arraybackslash}p{1cm}|
    >{\centering\arraybackslash}p{2cm} 
    >{\centering\arraybackslash}p{1.5cm} 
    >{\centering\arraybackslash}p{2.5cm}  
}
\hline
\textbf{Domain} & \textbf{Package} & \textbf{Lang.} & \textbf{Purpose}\\
\hline
\multirow{9}{1cm}{Symbolic} & jSymbolic \cite{jsymbolic} & java & analysis\\
& music21 \cite{music21} & python & analysis\\
& mido & python & MIDI processing\\
& pretty\_midi \cite{pretty_midi} & python & MIDI processing\\
& Humdrum \cite{humdrum} & awk & analysis\\
& Miditok \cite{miditok} & python & tokenize\\
& Muspy \cite{muspy} & python & generation\\
& note\_seq & python & MIDI processing\\
& Musicaiz \cite{musicaiz} & python & general\\
\hline
\multirow{2}{2cm}{Audio} & librosa \cite{librosa} & python & analysis\\
& madmom \cite{madmom} & python & general\\
\hline
\end{tabular}

\label{tab:softwares} 
\end{table}

%% file: 6-models.tex
\section{Models} \label{sec:models}
So far, we have seen the datasets that are commonly used in the music generation task and the software that can help us processing them. But what approaches and neural networks architectures have been used for generating music? Are the current AI-based models used in music tasks because they represent and learn correctly the music information or just because they perform good in other tasks? In this section we will introduce the architectures and approaches that are used in the music generation field.

Music generation with machines started around 1950s. Machines and projects like ILLIAC Suite in 1957 \cite{hiller1957musical} \cite{hiller1964computer}, Project1 by Koening in 1964, EMI by David Cope in the 1980s \cite{cope_emi}, \cite{cope2000algorithmic} and Analogiques A and B by Iannis Xenakis \cite{xenakis1981musiques}, \cite{harley2002electroacoustic}. The first AI-based models that used neural networks in the late 1980s where inspired by Algorithmic Composition \cite{nierhaus2009algorithmic}. Lewis in 1988 \cite{lewis1988creation} used a training and creation phases with a gradient descent approach. Also in 1988 and 1989, Todd \cite{todd1988sequential}, \cite{Todd1989} defined a sequential network for music applications. The sequential network used memory with feedback connections to the notes that were already produced. In 1994, Mozer \cite{Mozer94} used the CONCERT network by Elman that continued a sequence of notes given the probability over the possible note candidates. Apart from these previous works, there have been proposed several works that combine different algorithmic approaches and neural network architectures for music generation. Inside the Algorithmic Composition field we can find Markov Models, Generative Grammars, Cellular Automata, Genetic Algorithms, Transition Networks or Caos Theory \cite{hiller1957musical}. These models can compose music in different styles and ways, but new deep learning approaches are better for generalization than these rule-based approaches.

\subsection{Model Architecture}
Attending to the neural network architecture that is chosen for each model, we have two big groups of approaches: sequence models and generative models.
In Table \ref{tab:models} we provide a summary of some of the state-of-the-art models for music generation. We can observe how models have been adding the multi-track feature and how the architectures that have been used were firstly RNNs and now models are based mostly on Transformers. We can also see how Reinforcement Learning (RL) has not been explored in depth in the field nor Graph Neural Networks (GNNs).

\begin{table*}[!t]
\caption{State-of-the-art models for music generation. Melody refers to monophonic melody whereas polyphony refers to polyphonic music for one instrument, i.e., the piano, or multiple instruments.}
    \centering
    \begin{tabular}{
    p{2.8cm}  
    >{\centering\arraybackslash}p{2.2cm}
    >{\centering\arraybackslash}p{1.7cm}
    >{\centering\arraybackslash}p{2.5cm}
    >{\centering\arraybackslash}p{2cm} 
    >{\centering\arraybackslash}p{1.5cm}
    >{\centering\arraybackslash}p{3.5cm}
    }
    \toprule
        \multicolumn{7}{c}{Symbolic Music Generation}\\
        \midrule
         & Year & Architecture & Purpose & Instrument(s) & Multi-Track selection & Dataset\\
        \midrule
        Sequential \cite{Todd1989} & 1989  & Sequential NN & Melody & - & \xmark & - \\
        CONCERT \cite{Mozer94} & CS 1994 & RNN & Melody and chords & Piano & \xmark & JSB and Folk\\
        Bi-Axial LSTM \cite{EckS02} & ICANN 2002 & LSTM & Polyphony & - & \xmark & Blues\\
        RNN-RBM \cite{LewandowskiBV12} & ICML 2012 & RNN & Polyphony & 4 voices & \xmark & JSB Chorales\\
        VRAE \cite{vrae} & ICLR 2015 & VAE RNN & Melody & - & \xmark & Video Game songs\\
        BachBot \cite{liang2016bachbot} & 2016 & LSTM & Polyphonic & 4 voices & \xmark & JSB Chorales\\
        Song Pi \cite{song_pi} & 2016 & RNN & Melody and chords & Piano and drums & \xmark & MIDI Man\\
        RL-Tuner \cite{JaquesGTE17} & ICLR W. 2017 & RL & Melody & Piano & \xmark & Nottingham\\
        MidiNet \cite{yang2017midinet} & 2017 & CNN-GAN & Polyphony & 128 MIDI progs. & \cmark & Pop from TheoryTab\\
        Unit sel. \cite{unitselection} & ICCC 2017 & AE & Melody & - & \xmark & Wikifonia database\\
        PerformanceRNN \cite{music_transformer} & 2017 & Transformer & Piano pieces & Piano & \xmark & MAESTRO\\
        DeepBach \cite{deepbach} & ICML 2017 & RNN & 4 voices chorales & 4 & \cmark & JSB Chorales\\
        SeqGAN \cite{seqgan} & AAAI 2017 & GAN & Polyphony & Piano & \xmark & Nottingham\\
        DeepJ \cite{deepj} & ICSC 2018 & Biaxial LSTM & Polyphony & Piano & \xmark & Piano MIDI\\
        MusicVAE \cite{musicvae} & ICML 2018 & VAE & 4, 8 or 16 bar trios & 3 & \cmark & LMD\\
         C-RNN-GAN \cite{Mogren16} & 2018 & RNN GAN & Polyphony & - & \xmark & from web\\
         CNN-GAN \cite{binary} & ISMIR 2018 & CNN GAN & Polyphony & multi & \cmark & LMD\\
        MT \cite{music_transformer} & ICLR 2019 & Transformer & Piano pieces & Piano & \xmark & MAESTRO, JSB Chorales\\
        MuseNet \cite{musenet} & 2019 & Transformer & Polyphony & Piano & \xmark & BitMIDI, MAESTRO\\
        RL-Duet \cite{rlduet} & AAAI 2020 & RL & Mono Accomp. & - & \xmark & music21 JSB Chorales \cite{music21}\\
        AnticipationRNN \cite{hadjeres2020anticipation} & NCA 2020 &  LSTM & Melody & single voices & \xmark & JSB Chorales\\
        Music FaderNets \cite{FaderNets} & ISMIR 2020 & Fader NN & Polyphony & Piano & \xmark & VGMIDI and MAESTRO\\
        TransformerVAE \cite{transformer_vae} & ICASSP 2020 & Transf. VAE & Melody & - & \xmark & Hooktheory\\
        PopMAG \cite{ren2020popmag} & MM 2020 & Transformer & Mulri-track & 128 MIDI progs. & \cmark & LMD\\
        PopMusicTransf. \cite{huang2020pop} & MM 2020 & Transformer & Polyphony & Piano & \xmark & Pop piano covers\\
        MuseMorphose \cite{muse_morphose} & 2021 & Transf. VAE & Polyphony & Piano & \xmark & e LPD-17-cleansed\\
        CWT \cite{hsiao2021compound} & AAAI 2021 & Transformer & Polyphony & Piano & \xmark & Pop piano from the internet\\
        SymphonyNet \cite{symphonynet} & ISMIR 2022 & Transformer & Symphony & 128 MIDI progs. & \cmark & Symphonies\\
        Theme Transf. \cite{shih2022theme} & IEEE Mult. 2022 & Transformer & Polyphony & Piano & \xmark & POP909\\
        Perceiver AR \cite{perceiver_ar} & 2022 & Transformer & Polyphony & Piano & \xmark & MAESTRO\\
        Museformer \cite{museformer} & NIPS 2022 & Transformer & Polyphony & 128 MIDI progs. & \cmark & LMD\\

        \bottomrule
        \multicolumn{7}{c}{Audio Generation}\\
         \midrule
         SampleRNN \cite{samplernn} & ICLR 2017 & RNN & General audio & - & - & Beethoven piano sonatas\\
         
         MelNet \cite{melnet} & 2019 & CNN & General audio & - & - & MAESTRO\\
         It's Raw \cite{GoelGDR22} & ICML 2022 & S4 blocks & General audio & - & - & Beethoven, YouTubeMix\\
         Wavenet \cite{wavenet} & ISCA 2016 & CNN & General audio & - & - & MagnaTagATune\\
         Jukebox \cite{jukebox} & 2020 & VQ-VAE & Multiple genres & - & - & Private\\
         Musika \cite{musika} & ISMIR 2022 & GAN & Fast Generation & - & - & MAESTRO, Techno\\
         AudioLM \cite{audiolm} &  2022 & Transformer & General audio & - & - & MAESTRO\\
        \bottomrule
    \end{tabular}
    \label{tab:models}
\end{table*}

\subsubsection{Generative Models}
Generative models are based on Variational Bayesian (VB) Methods. This statistical part of ML treats statistical inference problems as optimization problems. These models are used to approximate the posterior probability for unobserved variables. In this family, we can find Variational AutoEncoders (VAEs), Generative Adversarial Networks (GANs), Diffusion Models (DMs), etc.

\paragraph{Variational AutoEncoders (VAEs)}
The VAE model was introduced in 2013 by Kingma and Welling \cite{vae}. The goal of a VAE is to model the inputs as continuous probability distributions to then decode new datapoints from the learned distributions. For this purpose, the architecture is similar to an AutoEncoder but in the case of the VAE, the latent variables form a latent space of probability distributions with means and variances. The encoder $q_{\phi}(\mathbf{z}|\mathbf{x})$ and decoder  $p_{\theta}(\mathbf{x}|\mathbf{z})$ are probabilistic with learnable parameters $\phi$ and $\theta$ respectively.
In a VAE, the encoder approximates the true posterior distribution $p(\mathbf{z}|\mathbf{x})$ in VB Evidence Lower Bound (ELBO) which enables facing statistical inference problems as optimization problems. The posterior and likelihood approximations are parametrized by the decoder. In Eq. \ref{eq:elbo} we show the loss function of a VAE which corresponds to the ELBO optimization, and which is composed by two terms: the reconstruction term and the KL divergence.

\begin{equation}
\begin{split}
\mathcal{L}(\theta, \phi; \mathbf{x}^{(i)}) =
&
-D_{KL}(q_{\phi}(\mathbf{z}|\mathbf{x}^{(i)}) \vert \vert p_{\theta}(\mathbf{z})) 
\\ &
+ \mathbf{E}_{q_{\phi}(\mathbf{z}|\mathbf{x}^{(i)}}) \left[ log p_{\theta}(\mathbf{x}^{(i)}\mathbf{x}|\mathbf{z}) \right]
\end{split}
\label{eq:elbo}
\end{equation}

The posterior inference is done by minimizing the Kullback-Leibler (KL) divergence between the encoder or approximate posterior, and the true posterior by maximizing the Evidence Lower bound (ELBO). This, in other words, means that we will try to reconstruct the inputs by capturing the distribution of the input data.
In these models, the gradient is computed with the so-called reparametrization trick. There are variations of the original VAE model
such as the $\beta$-VAE \cite{beta_vae} or the VQ-VAE \cite{van2017neural} which has been used for audio generation. An example of a symbolic music generation model based
on a VAE is MusicVAE \cite{musicvae} and an audio generation model based on a VQ-VAE is Jukebox \cite{jukebox}.

\paragraph{Generative Adversarial Networks (GANs)}
Generative Adversarial Networks (GANs) were introduced by Goodfellow et al. in 2014 \cite{gan}. These models are composed by two neural networks that are trained following the two-player minimax game (Eq. \ref{eq:gan}). These networks are called the \textit{generator (G)} and the \textit{discriminator (D)}. The generator creates data to try to fool the discriminator during training, and the discriminator learns to distinguish the real from the fake data generated by the generator.
The first multi-track model for music generation, MuseGAN \cite{musegan}, uses a GAN network.

\begin{equation}
\mathbf{E}_{x~p_{data}(x)} \left[ log D(x) \right] + \mathbf{E}_{z~p_{z}(z)} \left[ log \left( 1-D(G(z)) \right) \right] 
    \label{eq:gan}
\end{equation}

where $x$ is the data, $z$ are the noise variables, $p_{z}(z)$ is the prior defined on the input noise variables, $D$ is the discriminator's and $G$ the generator's functions. $D(x)$ is defined as the probability that $x$ is the real data rather than the data sampled by the generator's distribution.

\begin{figure}
    \centering
    \includegraphics[width=\columnwidth]{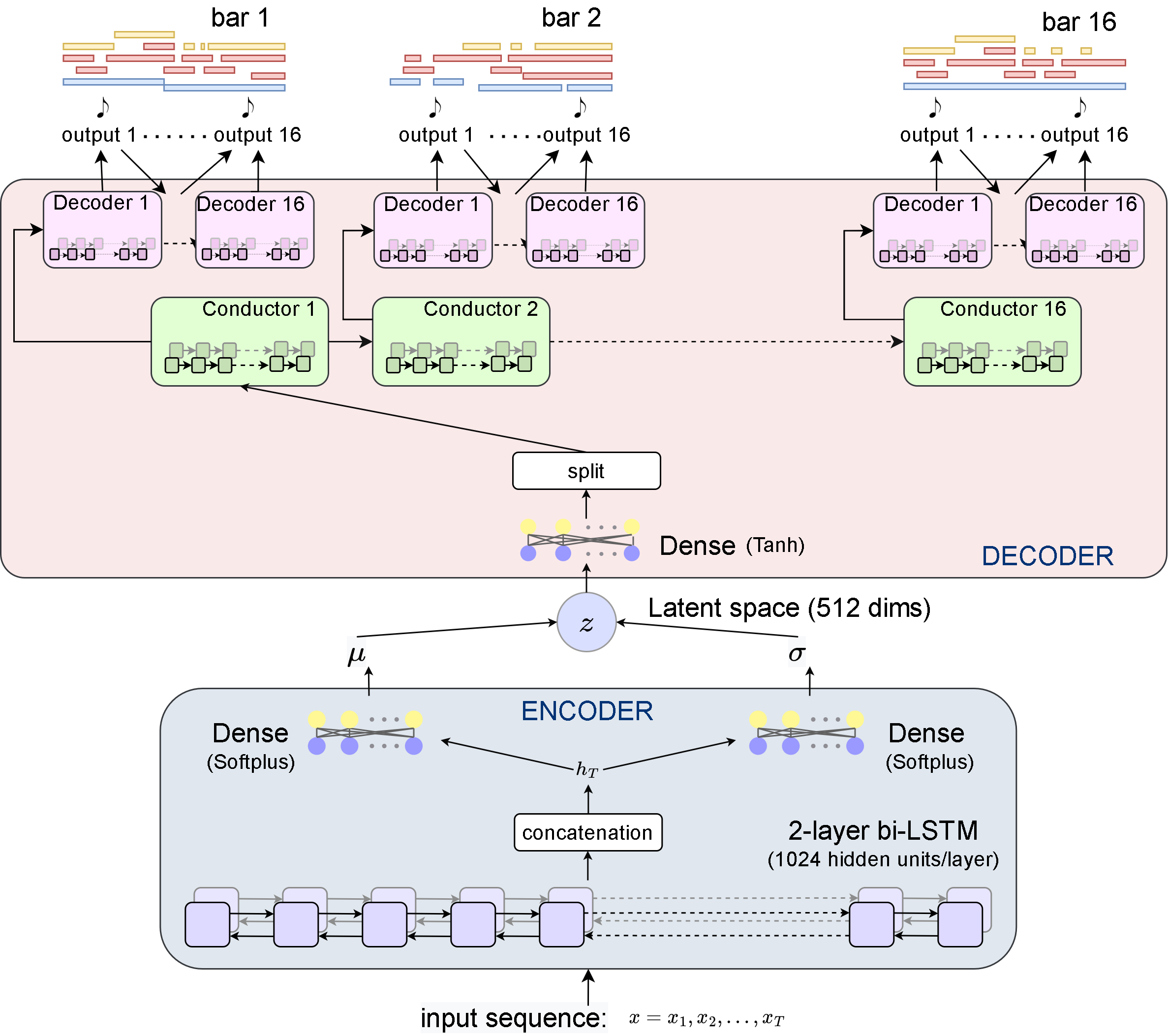}
    \caption{MusicVAE model architecture \cite{musicvae}. We can observe how the model presents an encoder with two BLSTM networks and a hierarchical decoder that generates the symbolic information note per note and bar by bar.}
    \label{fig:musicvae}
\end{figure}

\paragraph{Diffusion Models}
Diffusion models are inspired by non-equilibrium statistical physics \cite{sohl2015deep}. These models became popular in 2022 after the release of models like DALLE 2 \cite{dalle} or Stable Diffusion. However, as in most ML research, these models were firstly used in the field of Computer Vision. These models are based on the construction of a Markov chain of diffusion steps where a random noise is added in a forward process step by step to the data and then they learn to reverse the diffusion process to construct new data samples from the noise in a backward pass. Diffusion models learn with a fixed procedure and the latent variable has high dimensionality which is the same as the original data, and that makes them different from VAEs and GANs \cite{diffusionweng}. An example of a diffusion process applied to the music generation field is the work by Mittal et al. \cite{diffusion_models} that combined it with the MusicVAE to generate longer sequences.

There are more generative models and particular applications of them but as they have not been used in the music generation field we will not discuss them deeper. Examples of these models are the Flow-based models \cite{flows} which are characterized by constructing a sequence or \textit{flow} of invertible transformations to the data.

\subsubsection{Autoregressive Sequence Models}
Sequence models, in contrast with generative models, usually work with auto-regression. Auto-regressive models \cite{SchmidhuberH96}, \cite{Rosenfeld00}, \cite{BengioDVJ03} estimate the density of a sample with the chain rule of probability (Eq. \ref{eq:chainrule}).
\begin{equation}
    p(X) = \prod_{t=1}^{T} P(X_t \vert X_1,...,X_{t-1})
\label{eq:chainrule}
\end{equation}

where $X_t$ is a token of a sequence of $T$ tokens. The tokens are representations that depend on the domain (an audio frame, a pixel, etc.). Each new token can be obtained by estimating the conditional density on the prior tokens that are used to predict the $t$-th token. The tokens, in symbolic music, are note events such as the pitch, velocity or time delta.

\paragraph{Recurrent Neural Networks (RNNs)}
Recurrent Neural Netwroks (RNNs) are artificial neural networks that uses sequences or time series data.
These networks are characterized by the memory cells they have that allows to store information of modelling long-term sequences, but they suffer of vanishing gradient problems.
Some commonly used RNNs are Long short-term memory (LSTM) \cite{lstm} and Gated recurrent units (GRUs). In the music generation field, these were the first networks that were used to try to generate music with long-term dependencies. An example of that is the work by Eck and Schmidhuber in 2002  \cite{EckS02} that aimed to learn chords and melody from blues music.

\paragraph{Transformer-based Models}
Transformers were introduced in 1027 by Vaswani et al. \cite{attention}. These models have overtaken previous RNNs and have been used for a wide variety of purposes, from the NLP to the Computer Vision field. This is because the vanishing gradient that is a common problem in RNNs is avoided thanks to the attention mechanism which is the core of these models. The attention mechanism is based on previous works in retrieval systems and it is computed with three matrices: \textit{keys}, \textit{queries} and \textit{values}. The original Transformer \cite{attention} used the scaled-dot product attention or \textit{self-attention mechanism} (Eq. \ref{eq:att}).

\begin{equation} \label{eq:att}
    {\text{Attention}}(Q, K, V) = \text{softmax}\left(\frac{QK^{T}}{\sqrt{d_k}}\right)V
\end{equation}

where $Q \in \mathbb{R}^{d_{model}\times d_k}$, $V \in \mathbb{R}^{d_{model}\times d_v}$ and $K \in \mathbb{R}^{d_{model}\times d_k}$ are the queries, values and keys respectively and $d_{model}$ is the vocabulary size which, in symbolic music, can contain all the pitches, durations, tracks and any other features that we include in our encodings.

The original paper also includes an improved mechanism to improve the self-attention mechanism. Multi-Head attention allows to perform attention in parallel and attend information from different subspaces at different positions. In Eq. \ref{eq:multi-att} we show the general expression of Multi-Head Attention \cite{attention}. In this modification of the attention layer, $d_k = d_v = d_{model} / h$ where $h$ is the number of heads.

\begin{equation}
    \mathrm{MultiHead}(Q,K,V) = \mathrm{Concat}(h_1,...,h_n)W^0
    \label{eq:multi-att}
\end{equation}

where $h_i$ is the i$^{t}$ head attention: $h_i = \mathrm{Attention}(QW^Q_i, KW^K_i, VW^V_i)$.

The complexity of the self-attention mechanism is $\mathcal{O}(L^2 d)$ where $d$ is the layer size and $L$ the sequence length. This quadratic algorithm has been recently rethought and we now can find more efficient attention implementations that reduce the algorithm's complexity to linear \cite{wang2020linformer}, \cite{katharopoulos_et_al_2020}, \cite{VyasKF20}.

Transformer models that have been proposed in the recent years for NLP which have been trained with a large number of parameters can also be named as Large Language Models (LLMs). An example of these models are the GPT-based models \cite{gpt2}, BERT \cite{bert} or T5 \cite{t5}. Some of them uses the encoder-decoder proposed by the original Transformers and others such as the GPT only use decoders with casual attention. These LLMs have also been the core of some symbolic music generation models thanks to the relationship between text and music encodings and the pre-trained weights that we can use to perform \textit{transfer learning} from text to music.

\subsubsection{Reinforcement Learning} 
Reinforcement Learning (RL) is a ML method that is based on the achievement of a complex objective. Deep RL combines deep neural networks with RL to approximate the function that the agent needs to learn to interact with the environment. Thus, RL is learnt from interaction. This interaction occurs when the agent makes an action $a$ with a policy or set of rules to interact with the environment's state $s$ and gets a reward $r$ for it \cite{arulkumaran2017brief}. Each interaction defines a new state $s'$. The goal of the agent is to maximize the reward. We define $Q$ as the optimal function that gives the maximum future payoff for taking any action $a$ in a state $s$.  Deep Q-learning models the $Q$ function with deep neural networks.
An example of a RL model to generate music is the RL Tuner \cite{JaquesGTE17}. In this model, a LSTM network learns from the \textit{teacher} which is a RL algorithm that contains music theory concepts such as the scale. This allows the network to learn non-differentiable reward functions. This is a good approach for controlling the generated music on the music principles, however, there have not been lots of works in this area and RL for music generation still needs to be studied in depth. Another use of RL systems is to generate an accompaniment given a melody as is proposed in RL-Duet \cite{rlduet}.

\subsection{Input Representation}
Music representations vary depending on the domain, purpose and architecture of the model we aim to build. Both in the symbolic and audio domains we can find 1D and 2D representations of the input data.

\subsubsection{Symbolic}
Symbolic music information is usually structured in different levels: the piece-level, the track or instrument-level and the bar-level. This organization of music not only help training DL models but it also allows the evaluation, inpainting and communication in real applications.

\paragraph{File formats} We can find multiple file formats that store symbolic music. The most common ones are MIDIs (Musical Instrument Digital Interface) since they can be easily modified by music producers and composers in a Digital Audio Workspace (DAW). MIDI files contain the information as events, but they do not have all the music information such as chords or performance attributes such as dynamics \cite{oore2020time}. The most used events to encode symbolic music from MIDIs are the \textit{note on} and \textit{note off} (where the notes start and end), the \textit{pitch} (the value of the note from 0 to 128), the \textit{velocity} that stores the performance attribute of how much pressure is applied to the notes and the \textit{program number} which represents the instrument that contains the note (also from 0 to 128). Other MIDI messages are the \textit{tempo} or \textit{beats per minute} (bpm) changes, and the \textit{key} changes.
The most-used datasets for training DL models contain MIDI files. Other symbolic formats are the MusicXML which is richer that MIDI in what concerns to music attributes because it can contain harmonic information such as chords. The ABC notation is also a common used music notation. When it comes to develop APIs for music generation, we can send the music information is JSON formats or protocol buffers (protobufs). The \texttt{note\_seq} and \texttt{musicaiz} libraries can export music in such representaitons, and \texttt{muspy} only in the JSON format.

\paragraph{One-Hot Encoding} This encoding refers to a 2D matrix with pitch-time dimensions where matrix values are binary. The value is equal to 1 if at a certain time step (column), the pitch (row) corresponds to any of the pitches that are being played in the time step, and 0 if not.

\paragraph{Pianoroll} A pianoroll is a representation of note events that can belong to certain instruments. 

\paragraph{Sequence Encodings} When training sequence models it is necessary to map the input music to tokens. These tokens, as in NLP represent the music as a sequence of events in a 1D vector. In this encondings we find information about the notes, the structure, instruments, etc. There are different ways of encoding symbolic music and approaches to express the time delta units. PerformanceRNN \cite{performance-rnn-2017} in 2017 was the first model to introduce MIDI-like events as data structures. This encoding has been adopted for many works, and the recent Transformer-based models use this type of data structure but encoding the MIDI data in different ways. The Music Transformer \cite{music_transformer} was trained with note and time-events. Note events refers to measure time deltas or note durations in units of note lengths. For example, if our unit is the 16th note and we want to express the duration of a quarter note in deltas, we will have a time delta of 8 since there are 8 16th notes in a quarter note. Time-events do not use note lengths but time in milliseconds. Sometimes due to the nature or source of the MIDI data it is not quantized and selecting a note length as a unit can lead to approximations that would make us loose some groove of the original files. Music Transformer proposes to use time deltas as units of 10ms that leads to good results and respects the nature of the training data. In contrast, the Multi-track Music Machine (MMM) \cite{ens2020mmm} proposes an time encoding in symbolic note lengths instead of milliseconds. This encoding is based on LakhNES and improves its token representation by concatenating multiple tracks into a single lineal sequence with time shifts to define the time between MIDI events. This model uses a MultiInstrument and a BarFill representations. The MultiInstrument representation contains the tokens that the MMM model uses for generating multi-track music and the BarFill representation is used to inpaint bars, that is, replacing an existing bar for a new one that the model generates by maintaining the music principles coherence of previous and next bars. This was an important step to build models with which users can interact and that are close to score-based representations.
The Piano Inpainting Application (PIA) \cite{hadjeres2021piano} is similar to the MMM encoding since it also presents a MIDI-like encoding but, in spite of defining time shifts between events it tokenizes the duration of the notes, so the MMM's \texttt{NOTE\_OFF} token is not present in this encoding and the \texttt{NOTE\_ON} and \texttt{TIME\_DELTA} are replaced by the \texttt{PITCH} and the \texttt{DURATION} tokens, respectively.
The Compound Word introduced in the Compound Word Transformer (CWT) \cite{hsiao2021compound} groups tokens by their music properties as compound words, that is, a Note will be defined by its properties or attributes that are the pitch, duration and velocity tokens whereas positional tokens such as the Bar will only have the Bar tokens that indicates where a bar starts.

\begin{figure}
    \centering
    \includegraphics[width=\columnwidth]{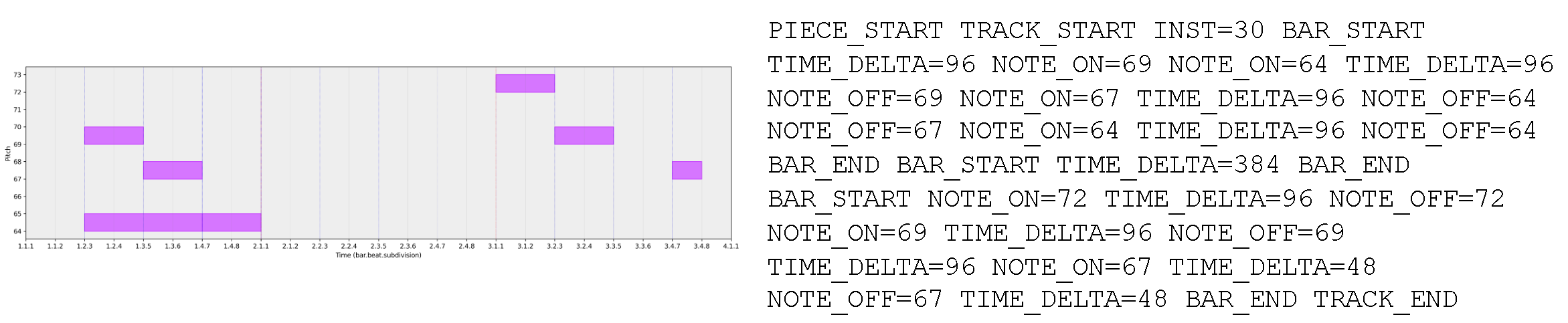}
    \caption{Example of a pianoroll representation with its MMM Track encoging. Time deltas are expressed in ticks in this example. We can observe how the piece, the track and the first bar start with the \texttt{PIECE\_START},  \texttt{TRACK\_START} and \texttt{BAR\_START} tokens respectively, and how the notes are opened with the \texttt{NOTE\_ON} and closed with the \texttt{NOTE\_OFF} tokens with the note's pitch as the value. The \texttt{TIME\_DELTA} is expressed in ticks in this example.}
    \label{fig:tokenexample}
\end{figure}

\begin{figure*}[!h]
 \centerline{
 \includegraphics[width=.75\textwidth]{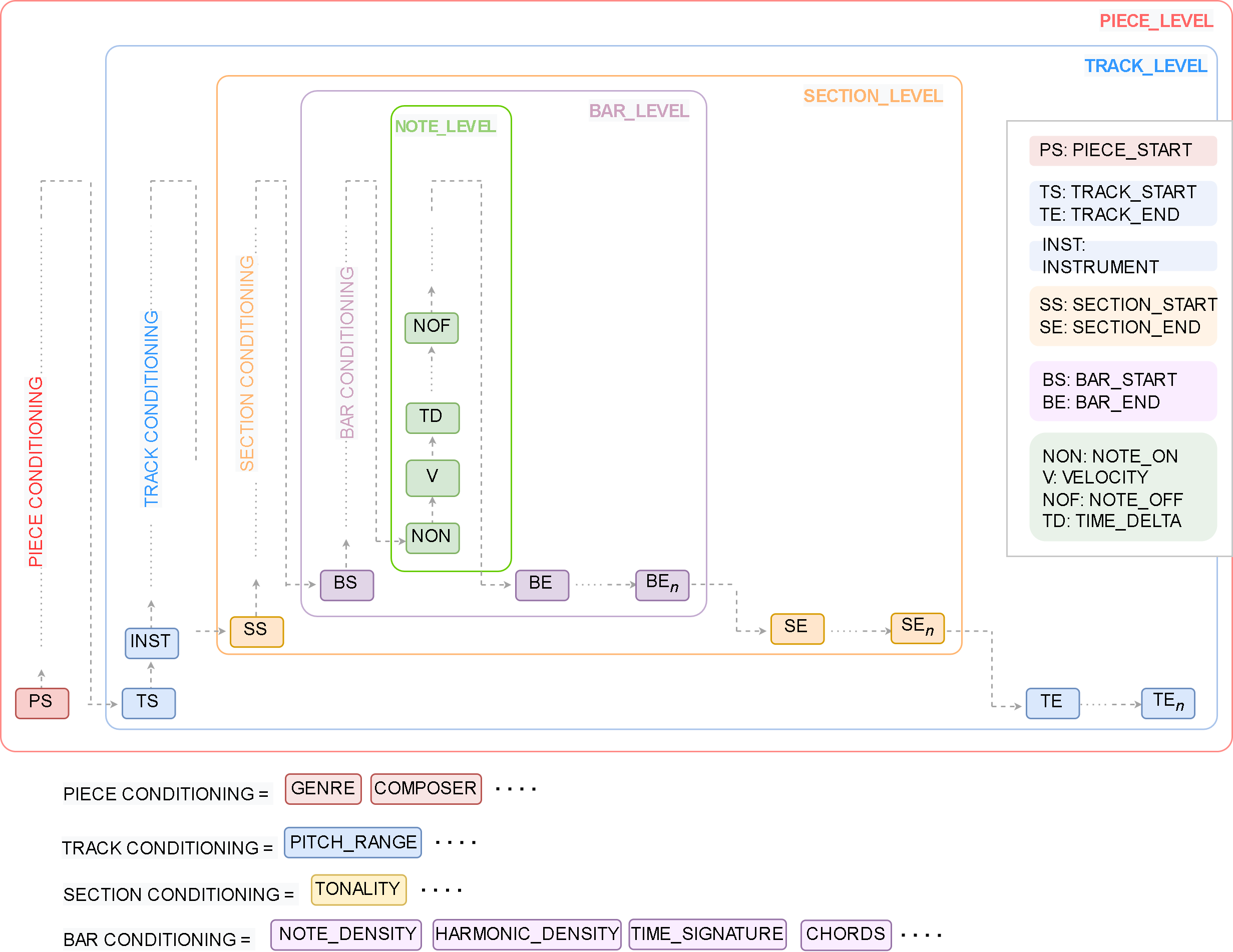}}
 \caption{A general encoding based on the MMM. The figure shows a general scheme configuration of how the MMM encoding could be extended for conditioning the generation at different levels. We extend the three levels proposed in the original MMM by adding one extra level that would require feature extraction algorithms on music structure detection or structure annotations. The levels are: the piece, instrument, section, bar and note levels. Adding the section-level tokens could allow to inpaint music phrases, themes or sections without the need of the user specifying or selecting them manually.}
 \label{fig:tokens}
\end{figure*}

In Table \ref{tab:tokens} we provide a summary of the different encodings for training sequence models, in Fig. \ref{fig:tokenexample} we show and example of the MMM tokenization extracted with \texttt{musicaiz} and in Fig. \ref{fig:tokens} we show a general schema of a symbolic music encoding. Note that it would be need a symbolic Music Structure Analysis (MSA) to add the section tokens that could be based on \texttt{SIA}, \texttt{SIATEC} and \texttt{COSIATEC} algorithms \cite{meredith2002algorithms}.

\begin{table*}[!t]
\caption{Encodings used in music generation state-of-the-art sequence models.}
    \centering
    \begin{tabular}{
    p{3cm}  
    >{\centering\arraybackslash}p{1.5cm}
    >{\centering\arraybackslash}p{0.5cm}
    >{\centering\arraybackslash}p{0.7cm}
    >{\centering\arraybackslash}p{1.5cm} 
    >{\centering\arraybackslash}p{1.5cm}
    >{\centering\arraybackslash}p{1.4cm}
    >{\centering\arraybackslash}p{1.1cm}
    >{\centering\arraybackslash}p{2.7cm}
    }
    \toprule
         & Time Sig. & Key & Chords & n. Program & Multi-Track & Bar Tokens & Inpainting & Time Unit\\
        \midrule
        MT \cite{music_transformer} & \xmark & \xmark & \xmark & \xmark & \xmark & \xmark & \xmark & 10ms\\
        Music BPE - MMR \cite{symphonynet} & \xmark & \xmark & \cmark & \cmark & \cmark & \cmark & \xmark & 32th note\\
        REMI \cite{huang2020pop} & \xmark & \xmark & \cmark & \cmark & \cmark & \cmark & \xmark & 32th note\\
        Octuple \cite{zeng2021musicbert} & \cmark & \xmark & \xmark & \cmark & \cmark & \cmark & \xmark & 64th to whole note\\
        MuMIDI \cite{popmag} & \cmark & \xmark & \xmark & \cmark & \cmark & \cmark & \xmark & 32 timesteps\\
        MMM \cite{ens2020mmm} & \xmark & \xmark & \xmark & \cmark & \cmark & \cmark & \cmark & 36th note\\
        CW \cite{hsiao2021compound} & \xmark & \xmark & \cmark & \cmark & \cmark & \cmark & \cmark & 16th note\\
        PIA \cite{hadjeres2021piano} & \xmark & \xmark & \xmark & \xmark & \xmark & \xmark & \cmark & 0-20s\\

         \bottomrule
    \end{tabular}
    \label{tab:tokens}
\end{table*}

\begin{figure}[!h]
    \centering
    \includegraphics[width=\columnwidth]{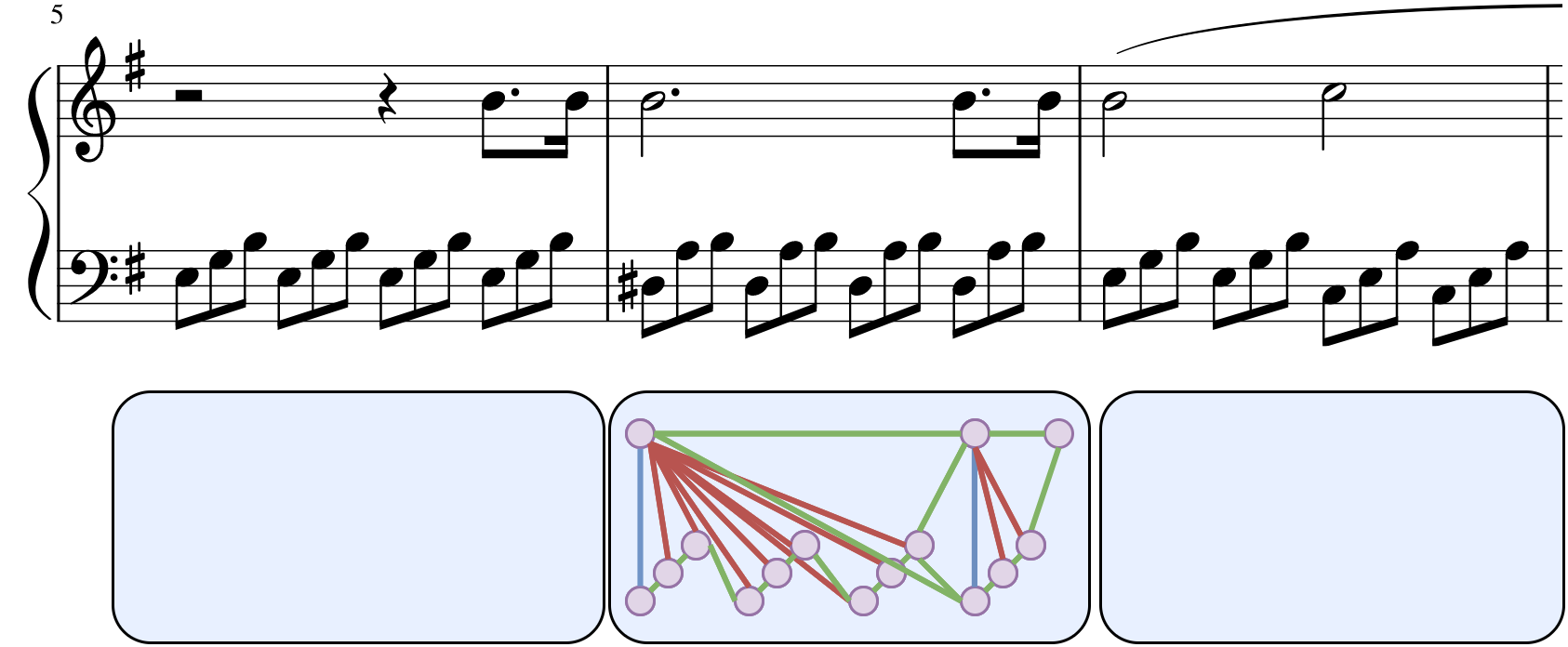}
    \caption{Graph representation of music proposed by Jeong et al. \cite{jeong2019graph}.}
    \label{fig:graph}
\end{figure}

\paragraph{Graphs} Graph Neural Networks (GNNs) \cite{gnn} are recent architectures that require to represent the input as graphs $G(V, E)$, where $E$ are the edges and $V$ the nodes. In the music generation field, this architecture has not been explored in depth yet. Jeong et al. \cite{jeong2019graph} proposed the first note-level graph representation of music as in which nodes represent the notes and edges the music relationship between the notes. Consecutive notes have different edges than notes that are played at the same time (chords). In Fig. \ref{fig:graph} we show an example of the music graph representation proposed by Jeong et al. This representation can be rethought and extended to add a bar-level, track-level or piece-level graphs to train further GNNs for music generation. Note that this representation works for MusicXML files where the note's time attributes are perfectly aligned with the beats and subdivisions in the bars, so in order to adapt this representation to MIDI files it would be necessary to quantize or modify the proposed representation.

\subsubsection{Audio}
\paragraph{Waveforms} A waveform is a 1D representation of the raw audio data as the amplitude of a signal in time. An important parameter to mention in waveforms is the sample rate or sampling frequency which is the number of samples per second in the signal.

\paragraph{Spectrograms} The spectrogram is the most-common input in generative models for audio. The spectrogram is a 2D representation of the spectrum on each segment obtained with the Short-Time Fourier Transform (STFT) which is computed with the Fast Fourier Transform (FFT). Normally, the Mel spectrogram is used which is nothing but a spectrogram where the frequencies are adapted to the mel scale. It is also common to use the log-mel spectrogram to logarithmically render the frequencies above a certain threshold so the frequencies are better scaled that in linear spectrograms. 

There are other transforms that could be used in music generation but its use is not commonly adopted. One of the reasons is that in VAEs, we need to reconstruct the input and then perform the inverse of the spectrogram or transform to get the amplitude and phase, which is not possible a priori in transforms such as the Constant Q-Transform (CQT). There are algorithms that invert these transforms but they are not mathematically exact. The Wavelets are invertible but due to their high computational cost, they are not common transforms used in the music generation field. In addition, some of these transforms have a high computational cost in comparison with FFT algorithm which is more efficient than most of other transforms.

\subsubsection{Data Augmentation}
It consists on modifying the input data to get more samples to train the models. The techniques that can be applied to the input data depend on the domain. In the symbolic domain we can modify the music data by transposing the key of the pieces. That is, adding or subtracting the same semitones value to all the pitches in the piece. Other possible data augmentation technique would be modifying the chord progression by substituting some chords by others. This would require having the annotations of the current chords in the piece (or predicting them), and make the replacement coherent so the final harmonic progression is coherent with the new chords. We should be aware of these modifications because we might loose the style of our pieces. As an example, if we condition a model to generate a Beethoven's piece, if we have augmented Beethoven's pieces to other keys the results the model might output would not be in the harmonic progressions or keys we might expect. However, we always can modify the generated music programatically to make it follow some rules or condition the DL model to certain keys or chords.
In what concerns to the audio domain, the modifications are based on the nature of the input but not in musical aspects. This is because symbolic data is easier to modify since we have the information about the notes, instruments, etc, but in the audio domain we do not have this information (unless we predict some music features with pre-trained models). Thus, the most-common techniques that can be applied to audio music signals are \textit{pitch shifting}, \textit{time stretching}, \textit{noise addition} etc.

\subsection{Output Form}
Briot et al. \cite{briot2020deep} define the output nature of music as one of the ways of classifying the music generation models. In this case, since in the audio domain the output form of the generated music depends on the dataset we train the model with, we will focus on symbolic music generation where we can identify more easily the nature of the output in music terms. The output nature is related to the structure and texture principles of music depending on what we look at. Some music generation models are trained to generate piano pieces whereas other models propose multi-instrument generation. The instruments that are present in a piece might be determined by the model itself or by the end user. This classification is a structure-based classification since texture is a complete different concept. A texture-based classification would be distinguishing between monophonic or polyphonic generation (we can have the same texture in single instrument and multi-instrument pieces).

\subsubsection{Attending to texture}
\paragraph{Melody Generation and Accompaniment}
Melody generation was one of the first goals of music composition models. A melody is a sequence of notes that can be monophonic or polyphonic. The first models that composed melodies generated short-term notes sequences. With the advent of RNNs and semantic models there were proposed works based on unit selection \cite{unitselection}. RNNs played an important role in the development of models in this field. Since the work by Eck and Schmidhuber in 2002 \cite{EckS02}, more models made use of RNNs to compose music. Examples of those models are the MelodyRNN by Google Magenta in 2016 \cite{waite2016generating} or the AnticipationRNN \cite{abs-1709-06404} in 2017. These models were capable of generating melodies and also chords that followed the melodies, but they were not developed to model \textit{counterpoint}. Counterpoint is the technique to combine melodies by following certain rules. DeepBach \cite{deepbach} in 2017 was one of the first models to achieve that. The model was trained with the JSB Chorales dataset \cite{LewandowskiBV12} that contain pieces of 4 voices that are witten in counterpoint. The architecture of DeepBach was composed by an RNN and used Gibbs sampling which is a technique that obtains a sequence of observations which are approximated from a multivariate probability distribution, to write the notes of each voice. Also in 2017 COCONET \cite{counterpoint_conv} was proposed. The model was composed by a Neural Autoregressive Distribution Estimator (NADE) model combined with Gibbs sampling.

However, these models could not generate new or creative music ideas such as high quality motifs. With the advent of new generative models such as VAEs, the MusicVAE\footnote{\url{https://magenta.tensorflow.org/music-vae}, accessed October 2022} \cite{musicvae} was proposed by Magenta in 2018. The idea was to interpolate in the VAE latent space to compose short music fragments, from 2 to 16 bars. The model was trained with melodies and trios (drums, bass and melody) of the LMD. After the MusicVAE was proposed, the music generation field aimed to generate longer sequences or melodies guided by the excellent capabilities of the new Transformer-based models in text generation. The Music Transformer \cite{music_transformer} proposed in 2018 with its \textit{screwing attention} mechanism was one of the first models that used an attention mechanism to generate polyphonic music. The model was trained with both JSB Chorales and the MAESTRO dataset for virtuosic piano. More models based on Transformers came up in 2018 such as MuseNet by OpenAI \cite{musenet}. The PopMAG \cite{ren2020popmag} published in 2020 can also generate accompaniments
The long-term dependencies were partially solved for a music sequence of 16 to 32 bars, that is, a music section inside a longer piece. The reason why longer sequences could not remember the previous music ideas was on the attention bottleneck when it comes to the context. From 2020 there have been proposed models that use both VAEs and Transformers to generate novel music and model long-term contexts. The PianoTree \cite{pianotree_vae}, the TransformerVAE \cite{transformer_vae} are examples of such models. However, any of these models have achieved the goal of generating a long music piece with a high-level structure coherence. One example of a model that uses diffusion is the model proposed by Mittal et al. \cite{diffusion_models} which is based in Denoising Diffusion Probabilistic Models (DDPMs) \cite{ddpm}. The model capture the temporal relationships among the VAE latents $z_k$ with $k = 32$ of a MusicVAE 2-bar model. This allows to extend the MusicVAE 2 bars to 64. However, as happens with the attention-based mdoels, this approach cannot follow a motif or music idea nor continue it with coherence. New attention-based models such as the Perceiver AR \cite{perceiver_ar} that uses cross attention to use a context of 4096 tokens is being trained with the MAESTRO dataset and seem to learn and generate longer music sequences.

Another approach to melody generation is to generate first a chord sequence and then a melody that matches the chord progression. This is a way of composing that humans also adopt. BebopNet model \cite{bebopnet} generates a melody from jazz chords because jazz harmony is more complex that other music genres. Other models use Variational AutoEncoders (VAEs) \cite{TengZG17}, Generative Adversarial Networks or GAN-based models \cite{jazzgan}, \cite{bebopnet}\footnote{\url{https://shunithaviv.github.io/bebopnet/}, accessed September 2022} and end-to-end models \cite{ZhuLYQLZZWXC18}. ChordAL \cite{Tan19} generates a chord progression that is sent to a melody generator to then send the output to a music style processor. 

\paragraph{Structure}
Generating music that follows a high-level structure is an open problem in the music generation field. The difficult part of generating structured music is the high understanding level that the models need to achieve this \cite{Lattner_thesis}. More precisely, the model not only needs to learn how rhythm, melody and harmony are combined but it also needs to remember music events that happened many bars or minutes before in order to develop or modify them.
The sections in music might be modelled as templates or can be learned by the models. Each music genre has its own structure or sections labelling such as the chorus and verse for rock or pop, or the exposition, development and recapitulation in the sonata form. Sections can also be labelled with capital letters.
In the past, there have been models that proposed templates to follow for modelling high-level structures. Lattner et al. in 2018 \cite{Lattner_thesis} used self-similarity matrices \cite{foote1999visualizing} and a Convolutional Restricted Boltzmann Machine (C-RBM) to impose a structure to the generated music. Other works have tried to develop models that learn the structure of music by themselves \cite{chen2019effect}.
Although new DL models that use the attention mechanism are trying to generate longer sequences, to our knowledge there is still no models that can generate structured music. Models that use cross attention mechanisms or linear attention might be a solution for this purpose. The Perceiver AR \cite{perceiver_ar} which was trained with the MAESTRO dataset is a recent model that could be used with this purpose. In spite of that, it has not been evaluated in terms of the music principles which let us doubt about its capabilities of generating structured music.
Apart from trying to work in the model's architecture which is the most common approach, researchers could develop algorithms that detect the structure of symbolic data to add structure tokens to the input encoding as we propose in Fig. \ref{fig:tokens}. This would also help in what concerns to inpainting.

\paragraph{Melody Harmonization}
Melody harmonization is the process to find a suitable and coherent chord progression for an existing melody. We need to point out the fact that there are a large number of chords, since we can build them from a tonic (the 12 notes in the chromatic scale), they can be inverted in one, two, three or more inversions depending on the chord complexity (triad, seventh, etc), and can have different qualities (major, minor, diminished, augmented, etc). In addition, a chord cannot be followed by all the chords in a vocabulary since the chord progression must have a coherence that is defined by each chord tonal function. We have also to distinguish between melody harmonization or generating an accompaniment for multiple tracks, but this last term is more related to the orchestration in our opinion.
The first models that face this problem were based on Hidden Markov Models (HMMs) which where outperformed by the RNNs. Eck and Schmidhuber in 2002 \cite{EckS02} used an LSTM to learn melodies along with chords. More recently, in 2019 LSTMs were also used to predict chord accompaniments for given melodies \cite{yang2019clstms}. Also in 2019 the Bach Doodle \cite{HuangHRDWHH19} (which used Coconet \cite{counterpoint_conv}) was developed to generate accompaniments for a given melody in the style of Bach. There are more works that used LSTMs proposed in the recent years \cite{sun2021melody}.

Regarding the generation of multi-track accompaniment it might be seen as part of the orchestration process. The models that aimed to perform this task are GAN-based architectures that implement lead sheet arrangements. MICA (Multi-Instrument Co-Arrangement) in 2018 \cite{ZhuLYQLZZWXC18} and MSMICA in 2020 \cite{ZhuLYZZC20} are an example of multi-track accompaniment.

\paragraph{Generation by Conditioning}
An important feature in music generation is the capability to condition a model. Conditioning can be based on make the model follow certain music principles or continuing a given prompt. Transformer-based models, thanks to the input encodings that researchers are proposing are being capable of generating music given a chord sequence \cite{symphonynet}. In this area we find a lot of research to be done due to the importance of the Human-Computer Interaction (HCI) research and to let users use this technologies by making them part of the composition process, for example, by inpainting at the different piece levels and difficulties that match the user's music knowledge.

\paragraph{Style Transfer} 
Music can be written in a wide variety of genres or styles. A genre is what allows to classify music. Different genres have its particularities and they combine the music principles differently. In the music generation field, style transfer refers to change the genre of a piece, for example, converting a rock song to a classical piece. This is an interesting technology due to the fact that there are no datasets for each music genre to train large DL models, however, it is also possible to perform transfer learning if our dataset is small \cite{ZhuangQDXZZXH21}. In the music generation field, Hung et al. proposed a recurrent VAE model for jazz generation \cite{HungWYW19} in 2019. 
In the DL field, style transfer was introduced in 2016 by Gatys et al. \cite{style_transfer}. The purpose was to apply style features to an image from another image. When it comes to music generation, the music style is changed by using an embedding or feature vector that represents the style to generate music. An example of a model that can perform this task is MIDI-VAE \cite{midi_vae} (2018). Thanks to the VAE's latent space, this model encodes the style in it as a combination of pitch, dynamics and instrument features that allows it to generate polyphonic music. As we mentioned before, the style of a music piece is defined by how the composer combine the music principles. Apart from MIDI-VAE which uses pitch and performance attributes to encode the style we can find other models such as MusAE \cite{musae} that also use pitch attributes to model the style of music. In addition to these models that used pitch attributes, we can also find models that use harmonic and texture attributes to encode the style, such as the PianoTreeVAE \cite{pianotree_vae} in 2020, MuSeMorphse \cite{muse_morphose} in 2021 and other works \cite{abs-2008-07122}.

\paragraph{Inpainting} With the new developments in Computer Vision that allow to combine the styles of 2 images or expand an existing painting, inpainting has become a need for end users. Inpainting is the ability to replace music content for novel content that matches the previous and next music segments maintaining the coherence of the piece. Examples of models that provide this capability in symbolic music are the Piano Inpainting Application (PIA) \cite{hadjeres2021piano} (2021) or the BarFill tokenization proposed in the MMM encoding \cite{ens2020mmm} (2020). PIA works for piano pieces while the MMM can handle multi-track music. The experiments on the models demonstrate how the replacement of existing music maintains the coherence of music aspects such as the harmony. However, with a better and more robust analysis and understanding of the symbolic music we could inpaint and control concrete music aspects of the music principles such as chords (the chord itself or its quality, inversion, etc), rhythmic executions (arpeggiated, maintained chords, etc) or high-level structure features such as the key of each section.

\subsubsection{Attending to instrumentation}
\paragraph{Music for Piano}
Music for piano is basically composing polyphonic music which has a melody and an accompaniment or chord sequence that is coherent with the melody. Researchers have focused on this type of music due to the lack of powerful models in the past that could handle multi-instrument music. From this perspective, \textit{music for piano} models are the same as the melody and accompaniment generation models that we classified inside \textit{attending to texture}. This is because piano is a polyphonic instrument with which we can play a melody and accompaniment at the same time, and the existing datasets for polyphonic music are, in its majority, for piano in exception of blues or jazz datasets.

\paragraph{Multi-instrument Music}
Multi-track or multi-instrument music\footnote{It is important to point out that when we talk about track or instrument, we refer to the notes that correspond to a certain instrument, which differs from the track concept in MIDI files.} refers to music written or arranged for more than one instrument.
Notice here that this classification differs from the texture because multi-instrument music can also be written as a melody with accompaniment or other textures. Music for multiple instruments adds an additional difficulty to music for piano due to the fact that there is more information (more notes) in the piece and because it adds the problem of handling each instrument symbolic and performance attributes to make a coherent piece. In Fig. \ref{fig:melody} we showed the scheme with the music basic principles of an output-like score for multiple instruments. 

The first approaches to multi-track generation started by including the drums track in the compositions. Since drums are non-pitched instruments, the generation only needs to match the rhythm of the melody. In 2012, Kang et al. \cite{semin2012automatic} proposed a model that could generate a chord accompaniment for a melody and also the drums. In 2017, Song of PI with its the hierarchical RNN proposed by Chu et al. \cite{song_pi} was also capable of generating the drums for pop music.
The first model to generate coherent music for multiple pitched instruments was MuseGAN \cite{musegan} in in 2017. From then, more GAN-based models such as SeqGAN \cite{seqgan} or CNN-GAN with binary neurons \cite{binary} were proposed. Later on, as happened with music for piano, Transformer-based models were developed for multi-track music. In 2019, Donahue et al. \cite{lahknes} proposed LakhNES, a model based on a TransformerXL and trained with the NES Music Database (NES-MDB). In 2020 more Transformer-based models with different encodings were also proposed for this task with (see Table \ref{tab:tokens}). Some of these models used its own encodings that we described in sequence encodings paragraph before in this section\footnote{Note that the encoding and the model are not the same thing. The encoding is how we structure the input data and the model refers to the neural network architecture that is trained}. Ens et al. prposed the MMM encoding \cite{ens2020mmm} that is trained with a GPT-2 pre-trained model on text by using Hugging Face Transformers library\footnote{\url{https://huggingface.co/docs/transformers/index}, accessed October 2022}. Other models that were released in 2020 are the PopMAG model which uses the MuMIDI encoding, MusicBERT that uses the OctupleMIDI encoding, Pop Music Transformer uses the REMI encoding and SymphonyNet uses the MMR encoding that is obtained with the Music BPE algorithm. Other work adress the pre-training of Transformer-based models for symbolic music understanding \cite{chou2021midibert}. It is important to notice that, since these models need large datasets to be trained in the case that we train them from scratch (more than 50K files if we look at the datasets we mentioned before), the computational cost of the encoding algorithms is important to accelerate the tokenizing process. It is also remarkable that reducing the number of tokens and using encodings that are not lineal but in a way where tokens can be coupled such as SymphonyNet does, makes training and inference faster.
It is worth to mention that, in spite that the Music Transformer's encoding \cite{music_transformer} was designed for piano pieces, it could be modified to encode multi-track music, or use the pre-trained model to perform transfer learning with multi-track datasets.

\paragraph{Instrumentation and Orchestration}
Instrumentation and Orchestration might not seem as relevant as melody composition but they play a fundamental role in the composition process.
However, there have not been a high interest on such techniques in the music generation field, probably because the models and datasets that are used in the field were not enough developed to face these problems.
As we introduced in Section \ref{sec:principles} instrumentation and orchestration are different techniques. The first one consists on the selection of instruments attending to their music and performance properties and the second one is the process of selecting and arranging the music information is the previously selected instruments \cite{sevsay2013cambridge}. Attending to the DL models that have tried to face and model these techniques, we can find SymphonyNet \cite{symphonynet}, which was proposed in 2022. It uses a BERT-based model and tokenization so that the model is trained to classify the instruments to which the notes belong. This is one of the firsts and successful ideas of music instrumentation and orchestration with DL models. However, it is important to notice that, for the model's perspective, it learns both techniques at the same time at it does not distinguish between them. As an example to explain this is that a piece for piano can be converted to a piece for an orchestra with these techniques, but SymphonyNet is trained directly with pieces for orchestra which means that it cannot convert piano pieces to orchestra nor reduce an orchestra piece for piano.

%% file: 7-hci.tex
\section{Human-Computer Interaction} \label{sec:hci}
As we mentioned in the introduction, the importance of human-computer interaction technologies is one of the key concepts of the entire music creation with AI. The co-creation has gained much importance that there are events and contests around the world such as the AI Song Contest\footnote{\url{https://www.aisongcontest.com/}, accessed October 2022} which that bring together teams that co-create music with AI. In this section we will dive into the user interfaces that have been proposed in the music generation field and the users. With this section we conclude the agents of the composition workflow that we showed in Fig. \ref{fig:general} to give way to the evaluation in the next section.

One of the most important elements in this HC interaction is the user interface. Research in modern user interfaces is increasing with recent advances of deep generative models and their new capabilities in the field of Computer Vision and this is fueling the birth of new interfaces for music generation with AI.
A well-designed interface not only gives the user a better experience but it also allows the user to interact at different levels with the model, especially in AI-based models for art \cite{LouieCHTC20}. In this part, the model is usually a black box that responds to the user prompts in the interface. Thus, there are two important agents in this part of the composition workflow: the user and the interface.

\subsection{Interfaces}
The interface is the first and only part that the user usually experiences with. Having a complex or basic interface are design decisions that depend on the tool or product that is aimed to build. The perfect interface does not exist, but a previous analysis of the target user and the knowledge about the technologies used to build a certain product can help to design a good interface.

Apollo by Bougueng and Ens (2019) \cite{tchemeube2019apollo} is designed to organize music data through the creation, editing,
and manipulation of musical corpora. Users can modify the music files they upload and generate music with parameters that control the note density, number of bars or pitch classes.

COSMIC (2021) \cite{Zhang2021COSMIC} proposes an interface for human-AI music co-creation with a chat that uses a NLP core that processes the human inputs as text and interacts with the human to define or refine the output music. The process starts with a start state in which the metadata is created, melody generation that matches the human's text inputs, melody revision with the HCI that allows to refine the melody with the given conditions, lyrics generation and revision also by interacting with the human and the end state when the session is closed. The music attributes that need to be passed to the first module of Natural Language Understanding are: Note Density, Pitch Variance, Rhythm Variance, Genre of Lyrics and the Keywords of Lyrics. These parameters are converted into keywords that are passed to the next module, the Dialogue State Tracker (DST). This is an example of a chat-like interface where the human's input is text.

Louie et al. (2022) \cite{LouieEH22} proposed two interfaces that aimed to measure the users interaction with two music generation models: the radio and the steering interfaces. Whereas the radio interface relies and a simple button that the user uses to generate music, the steering interface provides controls at different levels such as the pitch, key, etc, and it also allows the user to create a composition by selecting sequences that are made from the previous sequences. Louie et al. measure the user experience with the following indicators: the expression, communication, musical coherence, ownership, control and efficacy. In all these indicators, the steering interface obtained better results, however the user's level of experience or knowledge in music composition were not studied in depth which leaves an open question about if a more complete interface would be preferred by professional composers and a more simple interface would suit for amateur musicians, that is, if the interface has a direct relationship with the target user.

MusIAC by Guo et al. (2022) \cite{musiac} defines a control interface with track-level or bar-level controls which are based on the music structure. These levels allow to define controls that are related to the harmony or rhythm such as the tension. These controls based on the music principals are suitable for users with some music experience.

Other projects of SonyCSL such as NONOTO and PIANOTO developed in Javascript are proposing interfaces for inpainting symbolic music\footnote{\url{https://github.com/SonyCSLParis/music-inpainting-ts}, accessed October 2022}. While PIANOTO is based on a pianoroll-like interface where the user can inpaint by selecting zones in the interface and create symbolic music, NONOTO is oriented towards sounds creation and symbolic generation and inpainting with DeepBach.

The Magenta team has also developed plugins\footnote{\url{https://magenta.tensorflow.org/studio/ableton-live/}, accessed October 2022} for Ableton that use the models developed by the team in the back-end.

We saw different user interfaces to interact with an AI-based model for music generation with controls over the music principals in some cases, and other ``exploratory'' controls in other cases. We also saw differences from the number of controls that the interfaces have. There will be more interfaces proposed in the recent years and taking into account how text prompts and speech are succeeding in image generation applications, it is reasonable to assume that some of these prompt-like interfaces will also come for music generation, specially for beginner users that might just like to generate music by typing a sentence. An example for this would be the generation of music given the prompt: ``generate one minute music based on Chopin nocturnes in C major key, sad music''. In the audio generation domain, there is a recent model, AudioGen \cite{audiogen} that capable of generating audio given a text. In spite that the model do not generate music, it could be trained on music samples to achieve text guided music generation.

Now, we will dive into the users perspective and evaluation of the models when no interface is provided. A future research topic would rely on measuring the interface and user level all together to build more personalized interfaces depending on the users music knowledge or goal when using these technologies.

\subsection{Users}
The end user is the last agent in the composition workflow but, in our opinion, the most important one. The field of Computer Vision has been transferring knowledge from academia to industry for years. The feedback from end users is not only a key part in the industry level but also in academia. There are a few works that studied how users interact with music generation systems since the efforts in this field have been dedicated to the models and since the music generation field is new in comparison with Computer Vision there is a lack of feedback from users that could help the field grow in certain directions.

MuseGAN \cite{musegan} evaluates the model with subjective measures that are divided by the user's music knowledge. This is important because since music is subjective, the user's experience when interacting with a generative model for music generation can be biased. A good approach is to divide users by \textit{beginner}, \textit{intermediate} and \textit{pro} levels \cite{subjective}. These levels refers to users that have not music knowledge (beginner), users that have studied music (intermediate) and professional musicians or composers that can identify and analyze music in depth (pro). We think that taking into account the end users the field could propose different models with more or less controllability in the interfaces to better approach them.

%% file: 8-eval.tex
\section{Evaluation} \label{sec:eval}
Evaluating music generation is an open problem in MIR because there is a creative component in these systems that is hard to measure. In addition, since music is subjective, it is hard to evaluate or quantify the quality of a music piece. Researchers have proposed metrics for evaluating and comparing model performance, but a generalizable evaluation method for both symbolic and audio music generation is still lacking. Some of these metrics are common across different domains such as Computer Vision or Natural Language Processing fields such as the perplexity or the F-score, and other metrics are specific for the domain of study.
In the music generation field, before we evaluate we need to extract some features from the generated and ground truth music that is not always an easy problem. In the recent years, there have been proposed a wide-variety of measures that are divided into objective and subjective measures. It is also important to highlight that in the symbolic domain we can better measure musical aspects of the compositions whereas in the audio domain we might want to measure the quality of the generated signal. 

\subsection{Subjective Evaluation}
Subjective evaluation is related to the user's experience with the model. This kind of evaluation is done by users. The weak part of it is that each model is evaluated by different people, so the results are not reproducible. In addition, the user's knowledge, culture and fatigue can vary the results of the evaluation. Ji et al. \cite{survey} exposed the lack of correlation between the quantitative evaluation of music quality and human judgement which makes this evaluation necessary also because the end users are the ones that will interact and generate the music and their opinion is very valuable.

There are two common methods for evaluating music subjectively: the listening tests and the Turing test. Each model proposed in the AI field uses its own surveys and methods when doing listening tests, however there are recent works that propose surveys templates \cite{subjective}. An example of a listening test divided by user level is the one used in MuseGAN evaluation \cite{musegan} which was done by 144 users. More users took part of the evaluation of DeepBach \cite{deepbach} where 1.272 performed the listening test also grouped by their music knowledge. Normally, the listening tests consists on questions rated from 1 to 5 and the questions are aligned with the music principles. In the case of MuseGAN evaluation, the questions were made to measure the harmony complaisance, the rhythm, the structure, the coherence and the overall rating. We can let the users listen to the generated samples or provide also a score for a more in depth analysis.

Subjective evaluation can also be seen as a debugging tool for AI-based models. An example case is MusicVAE's listening test \cite{musicvae} which used Kruskal Wallis H-Test to validate the quality of the model, which conclude that the model performed better with the hierarchical decoder. This allowed researchers to verify which proposed architecture was generating better music, in spite that we should always be aware of the objective measures. 

We can not only ask users to rate the compositions or ask them to distinguish between a human or AI composition, but we can also ask non-quantitative questions related to the creativity of the model or the naturalness of the generated piece that is not possible with the objective evaluation. New studies on music generation evaluation \cite{chu2022empirical} compare DL models for music generation by asking the participants to measure the following criteria: \textit{overall}, \textit{melodiousness}, \textit{richness}, \textit{rhythmcity}, \textit{correctness}, \textit{structureness} and \textit{coherence} of the generated pieces in a 7-point Likert scale. Chu et al. \cite{chu2022empirical} prove that humans want more controllability in AI-based music generation systems and claim that \textit{melodiousness} is the most effective criterion to subjectively measure such systems. They define these criteria as:
\begin{itemize}
    \item Overall. The overall satisfaction with the music.
    \item Creativity. If the music is novel, valuable and original.
    \item Naturalness. If the music sounds like an expressive human performance.
    \item Melodiousness. How music and harmonious is the piece.
    \item Richness. How diverse and interesting is the music.
    \item Rhythmicity. Whether the rhythm is unified or not.
    \item Correctness. Whether the music has technical glitches such as sudden pauses, etc.
    \item Coherence. Whether the conditioned music is similar to the reference.
\end{itemize}

It is important to consider the domain we are working in when doing subjective studies since symbolic music needs to be synthesized and the synthesis process might affect the perception of the music in the users.

\subsection{Objective Evaluation}
In contrast with the subjective evaluation that can be similar for audio or symbolic music generation, the objective evaluation presents more differences due to the nature of both domains.
Before diving into specific methods for evaluating music generation systems, we will introduce the feature vectors that we can extract from symbolic music and some general measures that have been proposed in the literature.

\paragraph{Symbolic music} The feature vectors \cite{survey} that we can extract from symbolic music can be divided in the music principles: \textit{pitch-related}, \textit{rhythm-related} and \textit{harmony-related}.
\begin{itemize}
    \item Pitch vectors are the \textit{pitch range} that allows to measure the octaves in which a particular instrument or piece is composed and the \textit{intervals} which measures the pitch distance (normally in semitones) between consecutive pitches (we should consider both monophonic and polyphonic music here). We can also compute the \textit{pitch class histogram} (PCH) and the \textit{pitch class transition matrix} (PCTM). A pitch class (PC) is the pitch converted to a certain octave. This means that if we have a C4 and a C5, the PC will be the same for both since both of them correspond to the note C. Thus, we wil have 12 pitch classes corresponding to the 12 notes in a chromatic scale from C to B.
    \item Rhythm vectors are the \textit{inter onset intervals} (IOIs) which are the differences between the onsets of consecutive pitches, the \textit{note density} which is the number of notes in a music sequence, the \textit{duration range} which corresponds to the notes duration and the \textit{occupation rate} that measures how many steps of silence are in a music sequence. We can also compute the \textit{note length histogram} (NLH) and the \textit{note length transition matrix} (NLTM).
    \item Harmonic vectors are the \textit{polyphonic rate} which is the number of subdivisions where we find polyphony or the maximum number of polyphonic notes in a music sequence.
\end{itemize}

These vectors or descriptors allows not only to measure music but also to understand it, since most instruments can only be played in certain octaves and with particular rhythmic characteristics that the AI models should model.

We must take into account the fact that these features work for a music sequence which we can interpret at the piece-level (whole file or piece with all the instruments), track-level (only one instrument) and bar-level (a bar containing all the instruments or a bar of a certain instrument). The level that we choose in the evaluation would depend in the our goal.

Once we compute these vectors we can build evaluation measures and compare them with the training data. The measures are also divided in the music principles and are shown in Table \ref{tab:features}. The feature vectors, these measures can be calculated for each level: piece, track or bar-level.

\begin{table}[!t]
\caption{Common measures in symbolic music evaluation. Reproduced from \cite{ZhuangQDXZZXH21}.}
    \centering
    \begin{tabular}{
    p{0.8cm}  
    >{\arraybackslash}p{3cm}
    >{\arraybackslash}p{3.8cm}
    }
    \toprule
        & Measure & Definition\\
        \midrule
        \multirow{10}{1cm}{pitch} & Empty Bars & \% of empty bars\\
        & Pitch Class (PC) & number of PC\\
        & Qualified Notes (QN) & \% of notes $>$ 32nd note\\
        & Polyphonic Ratio (PR) & num of subdiv. with polyphony\\
        & Notes in Scale (NC) & \% of notes in the scale\\
        & Repetitions & num of same rhythmic seqs \\
        & Unique Notes (UN) & num of different PC\\
        & Consecutive Pitch Reps & num of same consecutive pitches\\
        & Tone Span (TS) & semitones difference between the highest and lowest pitches\\
        & Pitch Variations (PV) & num of different pitches\\
        \hline
        \multirow{6}{1cm}{rhythm} & Qualified Rhythm (QR) & \% of notes with a correct duration (quarter, 8th, 16th...)\\
        & Rhythm Variations (RV) & how many different durations do notes have\\
        & Pitch Duration Repetition (PDR) & pitch repetitions in a determined duration\\
        \hline
        \multirow{14}{1cm}{harmony} & Harmonic Consistency (HC) & measured with Impro-Visor \cite{keller2012impro}\\
        & Tonal Distance (TD) & semitones distance between tracks \\
        & Chord Histogram Entropy (CHE) & measures chord occurrences\\
        & Chord Coverage (CC) & num of chords in a music seq.\\
        & Chord Tonal Distance (CTD) & avg of TD between consecutive chords\\
        & Chord Tone to non-Chord Tone Ratio (CTnCTR) & measures how many notes do not belong to the current chord\\
        & Pitch Consonance Score (PCS) & measures how many notes are dissonant and consonant\\
        & Melody-Chord Tonal Distance (MCTD) & distances between melody notes and chord labels\\
         \bottomrule
    \end{tabular}
    \label{tab:features}
\end{table}

With more advances in harmonic extraction of symbolic music such as the HT model proposed by Cheng and Su \cite{chen2021attend} we could measure the chords and keys of the generated music and the samples in the train set, and propose new metrics that complement and expand the actual ones. This, obviously looking at the metrics to see the confidence of the extraction and thus, the metrics we could get from it.

Yang and Lerch (2020) \cite{yang2020evaluation} proposed a general objective method for evaluating symbolic music. First, some pitch and rhythmic features are extracted. Then, we build histograms from these features that we can approximate to Probability Density Functions (PDFs). From such functions we can measure the Kullback-Leibler Divergence (KLD) and the Overlapping Area to determine how similar the PDFs are. The PDFs are built from piece and bar levels and the comparison is done by cross-validating samples of the same dataset (\textit{intra set}) or confronting the training dataset with the generated dataset (\textit{inter set}).

\paragraph{Audio} In what concerns to measuring music audio signals, there are still not quantitative evaluation metrics to assess the overall compositional and musical quality of generated music \cite{musika}. The Frechét Audio Distance (FAD) allows to evaluate quantitatively the output of a raw music signal. Other ways of comparing state-of-the-art models are computing the log-likelihood on the test set \cite{perceiver_ar}.

%% file: 9-further.tex
\section{Further Perspectives and Research} \label{sec:future}
\subsection{Quantum Music Generation}
Eduardo Miranda in 2021 and 2022 introduced the Quantum Computing in Arts and Humanities \cite{miranda2021quantum}, \cite{miranda2022quantum}. Whereas quantum machine learning is a new research field for general applications, there are a few studies that are defining the first steps in the quantum music technology. This new MIR research field aims to develop tools to creating, performing, listening and distributing music  quantum computing tools.

\subsection{Emotions and Subjectivity}
Emotions in music have been widely studied \cite{yang2011music}. Since models that aim to detect emotions in audio signals to recent works that aim to generate music with certain emotions. There are specialized datasets for Music Emotion Recognition that could be used to train models to automatically label or condition  generative models.
In spite of that, emotions are subjective and they depend on the user's culture, background or music knowledge. 
An example of a music generation model with emotions is the one proposed by Bao and Sun \cite{bao2022generating}. The model uses a BERT-based model to recognize emotions (pre-trained model on GoEmotions dataset) and has three modules: an encoder-decoder architecture to generate lyrics and melodies, a music emotion classifier and a beam search algorithm that adapts the composition to the emotional conditioning.

\subsection{Copyright of the Generated Music}
A legal declaration about the copyright of the music obtained with an AI-based model is still in progress and it is a case of concern \cite{drott2021copyright} \cite{sturm2019artificial}. In the recent years, with the capital injection in companies that use AI-based tools and with the advent of new technologies such as Blockchain, there is a need of defining not only legal but also ethics statements to control AI-based music products \cite{morreale2021does}. There are several questions that should be addressed but one of the most relevant ones is: To whom does the copyright belong? There are different agents involved in the process of creating music with AI such as the people that made the database with which the models has been trained with, the team that builds and trains the model or the user that creates the music, among others. In spite of the fact that there are people who agrees that the tool or programmer should be considered in the copyright law, there are others that claims that the person who uses the tool should 
entitled to authorship since the
instantiation of an artwork is a sufficient condition for the attribution of authorship. However, if there is no human involved in the process of creating the music, according to the Designs and Patents Act 1988 (UK) (CDPA) the author should be the person that made the necessary arrangements for the creation of the work. According to Deltorn and Macrez \cite{deltorn2019authorship}, there are two approaches from the juridical perspective: \textit{de lege lata} and \textit{de lege ferenda}. The first one leads to deny the protection to computer-generated creations and analyze case-by-case to determine the human contribution. The second one relies on the necessity of modifying the existing copyright regime since software and databases (non human agents in the creation process) can be already protected with patents.

\subsection{Applications}
In spite that the tendency in the music generation field is nowadays in audio applications, music generation is gaining interest inside and out the MIR community. Not only the develop of music generation interfaces to explore and potentiate the HCI are emerging, but also commercial applications that allow users to generate and own their compositions. There are more applications of the music generation technologies that we have not covered in this paper although we show some of them below:

\begin{itemize}
    \item Music Education. With the upcoming technologies and recent studies in tasks like piano fingering, academia or industry could develop tools to generate music with annotations for the pianist to play a music phrase or segment. If AI-based models would have a good control over the harmony, rhythm, style or period, etc, these technologies could be also used as a composition learning systems for humans.
    \item Content generation. Content creators would benefit from these tools for generating new music for their projects. However, this is something that would require a strong legal analysis about the copyright of such creations. In spite that this would be more an industry application, it is closely related to how researcher can develop models with which humans could interact depending on their interests, goal or knowledge of music.
    \item Music Production. New deep learning-based music production technologies can benefit from music generation models for creating new layers or continuations of a given loop.
    \item Soundtracks. The generation of music for a video or film. This would be a step further from content generation due to the complex and long-term relationships and analysis that the models would require. Some of the things that the models should learn (or be conditioned with) would be the analysis of the characters, scenes, emotions, etc, that we want to evoke with the music by creating \textit{leitmotifs} \cite{kregor2017understanding}. This is not easy due to the fact that the networks should recognize characters and remember the motifs associated with each of them, apart from the emotional analysis which is subjective.
\end{itemize}

%% file: 10-discuss.tex
\section{Discussion} \label{sec:discussion}
We have seen in Section \ref{sec:human-ai} how music can be interpreted as a language from a technical perspective and we covered the relationship between the human and AI composition processes. In spite that human and AI ways of composing music are similar because both compose note by note or chord by chord, the understanding an perception of music between the human brain and AI approaches are far from being comparable. AI systems are designed to emulate human processes and we can relate the parts of the brain that understand language with LLMs as we saw in Section \ref{sec:human-ai}, but they do not work nor understand the world as a human brain does. Apart from that, humans have a creative component added to the previous knowledge that is still to prove in AI-based models. This creativity in AI would come with models that could extrapolate instead of interpolate. What seems to be true to us is the fact that AI-based models should be modelled to understand the music principles that we introduced in \ref{sec:principles}. This would allow to condition music generation systems and better model long-term time-frequency dependencies. From the domain knowledge perspective that we described in Section \ref{sec:audio_symb}, we saw how there have been a lot of work done in the symbolic domain. However, new studies seem to focus more on the audio domain also thanks to the research in speech technologies. Focusing on the data that we use to train our models which we describe in Section \ref{sec:datasets}, there is also a need of creating more datasets with high quality annotations for example with emotions, high-level structure, harmony or any other music principle that could serve for improving the current datasets. It is also important to measure the datasets biases since the majority of the research focus on the model. The model architectures that are described in Section \ref{sec:models} have been the key point of the research in the field and, in spite that we need new models, future research could focus on developing models that could be conditioned with the music principles. An alternative to large DL models that are being used at present would be the combination of Symbolic AI with these models, since music theory contains definitions that could be predefined so the model would not need to learn but to combine them. The conditioning would not a key part of the music composition process if it was not because of the interaction between the user and the machine that we exposed in Section \ref{sec:hci}. There is a lack of research done in the HCI in this filed if we compare it with the effort dedicated to the models design also because till now there have not been very promising results in music generation with AI. A future long-term line of research on interfaces could be the usage of brain-computer interfaces \cite{Tang2022} to create art. There is also less research done in the evaluation of generative models for music generation which is a very important part of the composition workflow. The \textit{quality} of the music is a wrong term when it comes to ``debugging'' a generated music sequence. There might be biases in each of the agents that we described in this paper that are important to analyze before we train our models and afterwards. We introduced in Section \ref{sec:eval} common metrics and ways to analyze subjectively and objectively the generate music, and there have been important advances in this line but there is still not a generalizable method that evaluates these systems at each agent level and in a more general perspective.

To sum up, the music generation field is growing and, as we introduced in Section \ref{sec:future}, there are applications and future research fields that could benefit from it that are still to appear. However, new models, better evaluation systems and more user interfaces could expand and let industry and academia learn new paths and interests that this field would explore.

%% file: 11-concl.tex
\section{Conclusions and Future Work} \label{sec:conclusions}
Music generation or composition is a research field that is growing with the advent of new models for arts generation. We have covered the agents that take part of the music composition process, we compared the human and AI processes to compose music and we discussed the applications that can benefit from this research field. We propose a few areas where the music generation field requires more research and attention: the creation of new datasets specially for the music genres that are not represented in the actual datasets, the analysis not only of the algorithms biases but also of the datasets biases, the development of more user interfaces, new research on conditioning the models on the music principles at different levels, research on the long-term modeling of music, develop more efficient encodings and models, and creating evaluation frameworks that would make research results reproducible. There is still room of improvement in what concerns to the long-term modeling of music which requires effort in both the models and music encodings designs. It is important to note that, although the field is moving towards long-term modeling, the generation of short sequences such as motifs is still unresolved and requires more attention. To our known, current models cannot generate high quality motifs and it is crucial to start from them and then develop and continue them to create complete phrases, themes and structured pieces.

For this purpose, neural network architectures that have not been explored in depth for music generation such as GNNs could be an interesting starting point. The encoding of music in a more efficient formulation would also allow to train deep learning models with less data and less training and inference times, better modeling long-term dependencies that are crucial in music generation, reducing the costs and impact in the environment. Generation by conditioning is also necessary to make the tools more flexible for users. This relies on having better analysis of the music in both audio and symbolic domains to be able to understand the music before we compose it, which this is something that humans do before creating music. From this side, there are algorithms or models of other MIR fields that can be already trained separately from the composition models and that could serve to condition the generative models. The integration of all these models or the creation of a multi-task or generalist model that would serve for both analysis and composition purposes could be a goal in this field, and developing new user interfaces would allow the users to co-create music in different ways and domains and extend the capabilities of the current technologies. Apart from that, understanding the human's brain from the neuroscience perspective is crucial to develop new AI technologies in what is called NeuroAI, and AI can also feedback neuroscience to help understanding cognitive processes. In either solution, domain or approach we take, we should always have in mind the genre or music style we are working with due to the intricacies and forms of each particular music genre.